\def\PREPRINT{}
\documentclass[10pt]{article}
\ifdefined\PREPRINT
  \usepackage[preprint]{tmlr}
\else
  \usepackage{tmlr}
\fi
\usepackage{amsmath,amssymb}
\usepackage{graphicx}
\usepackage{booktabs}
\usepackage{tabularx}
\usepackage{array}
\newcolumntype{C}[1]{>{\centering\arraybackslash}p{#1}}
\newcolumntype{L}[1]{>{\raggedright\arraybackslash}p{#1}}
\usepackage{pifont}
\usepackage[section]{placeins}
\usepackage{url}
\usepackage[hidelinks]{hyperref}
\title{Readable Is Not Steerable:\\Pre-Registered Tests of Physical Structure in Learned Fluid Representations}

\ifdefined\PREPRINT
  \author{\name Manvendra Modgil \email manvendramodgil.ai@gmail.com \\
          \addr Modint Intelligence}
  \hypersetup{pdftitle={Readable Is Not Steerable: Pre-Registered Tests of Physical Structure in Learned Fluid Representations}, pdfauthor={Manvendra Modgil}, pdfsubject={}, pdfkeywords={}, pdfcreator={}, pdfproducer={}}
  \newcommand{\ReleaseStatement}{The repository is public at \url{https://github.com/2025eb1100268-tech/readable-not-steerable}, and the scored checkpoints of Studies~2 and~3 are attached to its v1.0 release, so that readers can verify the committed hashes.}
  \newcommand{\AbstractRelease}{are public}
\else
  \author{}
  \hypersetup{pdftitle={Readable Is Not Steerable: Pre-Registered Tests of Physical Structure in Learned Fluid Representations}, pdfauthor={}, pdfsubject={}, pdfkeywords={}, pdfcreator={}, pdfproducer={}}
  \newcommand{\ReleaseStatement}{An anonymized copy of the repository is provided as supplementary material.}
  \newcommand{\AbstractRelease}{will be released}
\fi

\begin{document}

\maketitle

\begin{abstract}
Self-supervised world models are often described as organizing their latent spaces around physically meaningful quantities. We test this claim on two-dimensional decaying Navier--Stokes flow in three pre-registered studies, each tested once on held-out data opened a single time. In Study 1, a joint-embedding predictive architecture (JEPA) trained to predict the flow does not meet the pre-registered criteria for organizing its latent around fluid quantities better than an untrained network of the same architecture: the hypothesis is not supported. A JEPA trained on non-physical linear dynamics, and a static invariance model that never sees time, do as well or better. In Study 2, plain autoencoders that render the field make enstrophy prominent in their slowest unsupervised latent directions (best $|r|$ 0.80--0.86 against 0.38 for an untrained control, on fresh seeds and sealed data). Adding the VICReg regularizer lowers this prominence in every comparison, but not by the pre-registered margin for vorticity, while raising the autoencoders' linear decodability. The verdict is partially supported. In Study 3, we build steerable directions (``dials'') by routing information: the decoder can obtain overall scale only through a small conditioning code, and its other pathway carries only a degraded copy of spectral shape. Label-free rules find a loudness dial in all 8 trained models and a shape dial in 3 of 5 fresh seeds on sealed data, against a pre-registered bar of 4. Neither control shows a shape dial. The verdict is partially supported. Across the three studies, what a latent contains and what can be controlled through it come apart. In exploration, a supervised direction that reads spectral shape almost perfectly moved nothing when the latent was pushed along it. During the build, the loudness dial had to be identified by its effect on the output, as the direction closest to a pure rescaling, rather than by data variance; for shape, a variance rule and an effect-constrained rule each worked in 3 of 5 seeds. The full program used about 70 logged hours of compute on one consumer GPU; all pre-registrations, deviations, code and outputs \AbstractRelease{}.
\end{abstract}

\section{Introduction}\label{sec:1}

In our experiments on two-dimensional fluid flow, the latent structure that is usually read as a model ``organizing around physics'' depended more on the training objective, its regularizer and the decoder than on the dynamics the model was trained to predict. What a latent contains and what can be steered through it also turn out to be different properties. This paper reports three pre-registered studies that reach these conclusions, each tested once on held-out data opened a single time.

Self-supervised world models learn to predict how a scene evolves. Joint-embedding predictive architectures (JEPAs) predict in a learned latent space rather than in pixels (\citealp{lecun2022path}; \citealp{assran2023ijepa}; \citealp{bardes2024vjepa}). Work on intuitive physics (\citealp{piloto2022intuitive}) and on data-driven discovery of conservation laws (\citealp{liu2021conservation}) invites reading the latent structure of such models as evidence that they learn physical quantities.

That reading usually rests on two kinds of evidence. A linear probe decodes a quantity from the latent, or a principal or slow direction of the latent correlates with a conserved quantity. Neither is informative without controls. An untrained convolutional network already exposes many field statistics linearly, and a direction that correlates with a quantity need not be one along which the model represents or controls it.

We therefore separate three properties that are often conflated:

\begin{itemize}
\item \textbf{Decodability:} a quantity can be read from the latent by a linear probe.
\item \textbf{Prominence:} the quantity dominates the latent's main unsupervised directions, so it can be found without labels.
\item \textbf{Steerability:} moving the latent along a direction changes that quantity in the decoded output, and nothing else breaks.
\end{itemize}

All three are measured on two-dimensional decaying Navier--Stokes flow, where the relevant quantities (energy, enstrophy, spectral shape, vortex structure) can be computed exactly. Every study was pre-registered, evaluated once on a held-out test set, and compared against controls designed to break the mechanism being claimed. The controls are a JEPA trained on non-physical dynamics, a static model that never sees time, an untrained network, and a version of our final model without the key design ingredient.

Our contributions are:

\begin{enumerate}
\item \textbf{A pre-registered negative result (Study 1).} A JEPA trained to predict the flow beats an untrained network of the same architecture on three probe targets but fails the pre-registered criteria for organizing its latent around fluid quantities. A JEPA trained on non-physical linear dynamics reads enstrophy, palinstrophy and vortex count significantly better than the JEPA trained on Navier--Stokes, and a static VICReg model with no time beats it on all four structure targets (enstrophy, palinstrophy, vortex count and the Okubo--Weiss fraction).
\item \textbf{A mechanism (Study 2).} Plain autoencoders that render the field make enstrophy prominent in their slowest unsupervised directions, whether they render the present or the future. Adding VICReg lowers that prominence in every comparison while raising the autoencoders' decodability, so prominence and decodability dissociate. The pre-registered size of the drop held for one render target and not the other, so the verdict is partially supported.
\item \textbf{A design for steerable dials (Study 3).} A decoder conditioned through a small code yields a label-free loudness dial in every trained model, and a label-free spectral-shape dial in 3 of 5 fresh seeds on sealed data, against a pre-registered bar of 4. The shape dial forms only when the content pathway's copy of spectral shape is degraded, and never in either control. The loudness dial had to be found by its effect on the output rather than by data variance (E). In exploration, replacing sub-pixel upsampling with resize-convolution removed the grid artefacts that had disqualified strong dials (E).
\item \textbf{A complete record.} Every pre-registration, amendment, deviation and sealed-set opening is dated and tied to a commit. Every result is tagged as confirmatory, exploratory or post hoc (Section~\ref{sec:3.4}). The whole program used about 70 logged hours of compute on one consumer GPU.
\end{enumerate}

The scope is narrow on purpose. All results concern $64 \times 64$ periodic two-dimensional decaying flow, one simulator and initial-condition distribution, and a small convolutional encoder. None of the results, positive or negative, is claimed to transfer to three-dimensional flow, larger models or other data.

Section~\ref{sec:2} places the work in context. Section~\ref{sec:3} describes the data, the measures and the protocol. Sections~\ref{sec:4}--\ref{sec:6} report the three studies, Section~\ref{sec:7} discusses what transfers and what does not, and Section~\ref{sec:8} documents the process.

\section{Related work}\label{sec:2}

\textbf{Predictive world models.} JEPAs learn by predicting the latent representation of one part of the input from another, avoiding pixel reconstruction (\citealp{lecun2022path}; \citealp{assran2023ijepa}; \citealp{bardes2024vjepa}). Collapse is prevented by an exponential-moving-average target and, in our setting, by variance--covariance regularization in the style of VICReg (\citealp{bardes2022vicreg}). Self-supervised joint-embedding methods have also been applied to partial differential equations, using Lie point symmetries of the PDE as augmentations (\citealp{mialon2023lie}). Our static control is closest in spirit to that work: it learns invariance to symmetry transformations of a single frame and never sees time.

\textbf{Physics from data.} Neural simulators and operators learn to time-step physical systems (\citealp{sanchezgonzalez2020gns}; \citealp{li2021fno}). A separate line seeks physical structure in what models learn. That includes intuitive-physics concepts tested by violation of expectation (\citealp{piloto2022intuitive}), conserved quantities recovered from trajectories (\citealp{liu2021conservation}), and conservation built into the architecture as an inductive bias (\citealp{greydanus2019hnn}; \citealp{alet2021noether}). We do not propose a discovery method. We test whether a structure that such work would read as physical survives controls designed to produce the same structure without physics.

\textbf{Disentanglement and controllable generation.} Unsupervised disentanglement is impossible without inductive biases on both the models and the data (\citealp{locatello2019challenging}), and penalties such as the $\beta$-VAE objective (\citealp{higgins2017betavae}) supply one kind of bias. Our dial models supply a different one: they control which pathway can carry a given piece of information to the decoder. The dial code conditions the decoder through feature-wise affine modulation (FiLM; \citealp{perez2018film}).

\textbf{Upsampling artefacts.} Transposed convolutions produce checkerboard artefacts through uneven kernel overlap (\citealp{odena2016deconv}). Sub-pixel convolution (\citealp{shi2016subpixel}) is equivalent to a transposed convolution with stride and shares the tendency; an initialization that mimics nearest-neighbour resizing removes it at the start of training (\citealp{aitken2017checkerboard}). We find the same artefacts in the tails of strongly steered latents, where they decide whether a direction passes as a usable dial.

\textbf{Probing versus intervention.} A probe that decodes a property shows the information is present, not that the model uses it (\citealp{belinkov2022probing}). Control tasks calibrate how much of a probe's accuracy reflects the representation rather than the probe (\citealp{hewitt2019control}), and amnesic probing tests use by removing the property and measuring behaviour (\citealp{elazar2021amnesic}). Our untrained-network baselines play a role similar to control tasks, and our steering harness tests use by intervening on the latent and measuring the decoded output.

\textbf{Pre-registration in machine learning.} Pre-registered submissions, reviewed before results are known, were piloted at NeurIPS 2020 and 2021 (\citealp{bertinetto2021prereg}; \citealp{albanie2022prereg}). We follow a lighter protocol without prior review: hypotheses, rules and analysis code are committed before a held-out test set is opened once by a frozen script, and every later change is logged.

\section{Setting and protocol}\label{sec:3}

All three studies share one simulator, one family of encoders, three kinds of measurement and one protocol for confirmation. Details that differ between studies are given in Sections~\ref{sec:4}--\ref{sec:6}.

\subsection{Flow data}\label{sec:3.1}

We simulate two-dimensional incompressible Navier--Stokes flow in vorticity form on the periodic domain $[0, 2\pi]^2$, with a pseudo-spectral solver at $64 \times 64$ resolution, 2/3-rule dealiasing, integrating-factor (Lawson) fourth-order Runge--Kutta time stepping at CFL 0.3, and float64 arithmetic. Runs are unforced and decaying. Initial conditions are, in equal shares, periodic Gaussian vortex blobs (2--12 vortices of random sign and position, radius log-uniform in [0.25, 0.6]) or random fields with energy spectrum $E(k) \propto k^4 \exp(-2(k/k_0)^2)$, $k_0 \in [2, 5]$. Each field is scaled to a root-mean-square vorticity drawn log-uniformly from [2, 4], which sets the initial enstrophy $Z_0$. Viscosity $\nu$ is drawn log-uniformly from $3\times10^{-3}$ to $5\times10^{-2}$. A quarter of Study 1's trajectories are inviscid ($\nu = 0$); later sets raised the inviscid share to 378 of about 1,000, and the matched-energy sets are entirely inviscid. The lower bound on $\nu$ is the smallest value at which $64^2$ runs track $128^2$ runs (95th-percentile palinstrophy error 6\%).

Snapshots are stored every 0.04 time units, 64 per viscous trajectory. Inviscid trajectories are truncated at 24 snapshots, because at this resolution enstrophy otherwise accumulates in the finest resolved shells. The solver passes tests for inviscid conservation of energy and enstrophy, for the energy budget $dE/dt = -2\nu Z$, for two analytic solutions (Taylor--Green and Lamb--Oseen vortices), for a divergence-free velocity and for fourth-order convergence. A flipped sign in the advection term passes the conservation, budget and analytic-solution tests, so three more tests were added that catch it: an analytic right-hand side, the rotation direction of a vortex pair, and recovery of $\nu$ from stored frames.

Splits are made by trajectory, never by frame. Study 1 used 2,000 trajectories (1,295 train, 279 validation, 276 in-distribution test, 150 out-of-distribution test), where the out-of-distribution set is an interior Reynolds-number band ($\mathrm{Re}$ 286--389) held out of training and selection. Later stages used further sets from the same generator, which excluded the retired out-of-distribution band, raised the inviscid share, and were tested for disjoint seeds and initial-field hashes against Study 1's training and validation trajectories and against each other (the retired test splits were not read).

\subsection{Encoders}\label{sec:3.2}

In Studies 1 and 2, every model reads the velocity field $(u, v)$ as a two-channel image with one global input scale. The encoder is a small convolutional network with circular padding, GELU activations, Kaiming initialization and no normalization layers; it maps a frame to an $8 \times 8 \times 128$ feature map, whose spatial mean is the 128-dimensional pooled latent. We avoid per-sample normalization because, in planning pilots before Study 1 (summarized in the run log but not reproducible from the repository), instance normalization, or group and layer normalization with bias-free convolutions, reduced the log-energy probe $R^2$ to 0.01--0.02, removing a basic field statistic from every latent. In Studies 1 and 2 the untrained control uses exactly the initialization that trained models start from. Study 3's models use two encoders of this design and a different untrained control (Section~\ref{sec:6}).

\subsection{Three kinds of measurement}\label{sec:3.3}

The measures correspond to the three properties defined in Section~\ref{sec:1}.

\begin{table}[htbp]
\centering
\small
\begin{tabularx}{\linewidth}{l>{\raggedright\arraybackslash}Xl}
\toprule
Property & Measure & Used in \\
\midrule
Decodability & Ridge-regression probe from the $z$-scored latent to a target; $R^2$ on held-out trajectories & Studies 1--3 \\
Prominence & Correlation between a target and the latent's slowest unsupervised directions & Studies 1--2 \\
Steerability & Change in the decoded field when the latent is moved along a direction & Study 3 \\
\bottomrule
\end{tabularx}
\end{table}

\textbf{Targets.} From each true field we compute log energy $E$, log enstrophy $Z$, log palinstrophy $P$, log maximum $|\omega|$, log vorticity kurtosis, log dissipation $2\nu Z$ (viscous runs), $\log\nu$, log(1 + vortex count) and the Okubo--Weiss fraction. Vortex cores are regions where the Okubo--Weiss parameter $Q$ (\citealp{okubo1970}; \citealp{weiss1991}) is below $-0.2$ times its standard deviation, counted as connected regions across the periodic boundary with a minimum of 4 pixels; the Okubo--Weiss fraction is the vorticity-dominated area. In decaying flow many quantities fall with time, so the structure-probe scores (S1, Section~\ref{sec:4.2}) are reported after partialling out time crossed with initial-condition family (fitted on training data), so that a latent that merely encodes how late a frame is gets no credit.

\textbf{Prominence.} On inviscid trajectories, where energy and enstrophy are conserved, we $z$-score the pooled latent, reduce it to 32 principal components, and solve the generalized eigenproblem that finds directions varying least within a trajectory relative to across trajectories (with Ledoit--Wolf shrinkage on the within-trajectory covariance; \citealp{ledoit2004}). The criterion is related to slow feature analysis (\citealp{wiskott2002sfa}) and to linear discriminant analysis with trajectories as classes. The statistic, written E1 for enstrophy, is the largest $|r|$ between a trajectory's mean projection on one of the three slowest directions and its log initial enstrophy, averaged over seeds. Directions are fitted on training and validation trajectories and scored on held-out ones. A high E1 means enstrophy is prominent in the latent; it does not mean the model discovered or knows that enstrophy is conserved, and a latent can encode enstrophy well without making it prominent (Section~\ref{sec:5.3}).

\textbf{Steerability.} For a direction $d$ with training-set spread $\sigma_d$, we decode $z + \alpha\sigma_d d$ for $\alpha$ from $-2$ to $+2$ in steps of 0.5 and measure the decoded vorticity field. Two coordinates summarize it: amplitude $a = \log E$ (loudness) and shape $s = \log(Z/E)$, the log of the energy-weighted mean squared wavenumber. Rescaling a field changes $a$ and leaves $s$ fixed. The shape selectivity of a direction is $\sigma_{\mathrm{shape}} = |\Delta s| / (|\Delta a| + |\Delta s|)$, close to 0 for a loudness dial and close to 1 for a shape dial. A direction counts as a dial only if all of the following hold:

\begin{itemize}
\item its effect is at least 0.2 log-units per unit $\alpha$: the slopes of $a$ and $s$ against $\alpha$ are fitted per frame by least squares, and the effect is $|\mathrm{median}\,\Delta a| + |\mathrm{median}\,\Delta s|$ over frames. It must also be monotone: the median absolute Spearman correlation between $\alpha$ and the dominant coordinate ($a$ or $s$, whichever has the larger median slope) is at least 0.9;
\item re-encoding the steered field recovers the commanded move: the decoded vorticity is converted to velocity and re-encoded, and the change of its projection on $d$ has Pearson correlation of at least 0.9 with the commanded $\alpha\sigma_d$ over all (frame, $\alpha$) pairs, and a least-squares gain between 0.5 and 2. The change must also be frame-specific: at most half of the $\alpha = \pm 2$ change may be one pattern shared by all frames;
\item the steered fields stay realistic: the energy spectrum stays inside the training envelope no worse than an ideal dial of the same strength (tolerance $-0.05$) at every $\alpha$, and no more than 1\% of frames carry grid artefacts at any $\alpha$, defined as more than $10^{-2}$ of the field's power outside the solver's dealiasing band;
\item for a shape dial $\sigma_{\mathrm{shape}} \geq$ 0.7; for a loudness dial $\sigma_{\mathrm{shape}} \leq$ 0.2.
\end{itemize}

Every criterion must hold separately on each of the two scoring sets. The effect floor was raised from 0.1 to 0.2 after the first stage of Study 3, and the artefact criterion was calibrated on one known-bad case; both changes were committed before they were used to score anything (Appendix~\ref{sec:8.2}).

\subsection{Protocol and claim tags}\label{sec:3.4}

Each confirmatory study followed the same core steps: hypotheses, decision rules and analysis code were committed before the test data were opened, a frozen script opened the data once, and its outputs were committed as written. Studies 2 and 3 added two safeguards. Their test sets were sealed in code: the loader refused to open a set unless a flag set by the frozen evaluation script was present and the named pre-registration was committed and unmodified. The SHA-256 hash of every scored checkpoint was also committed before opening. Study 1's test sets were single-use by protocol rather than by code, its checkpoints were identified by run name, and its final configuration was committed and dry-run on training and validation data before the single test run. Before each exploratory stage, predictions with probabilities were committed and never edited afterwards.

Every result in this paper carries one of three tags. \textbf{Confirmatory (C)} means the rule was fixed in a committed pre-registration and evaluated once on sealed data; within C we distinguish primary (decides the verdict), secondary (rule stated in advance, not part of the verdict) and descriptive (pre-registered to be reported, with no rule). \textbf{Exploratory (E)} means the result came from exploration data in stages chosen adaptively. \textbf{Post hoc (PH)} means the analysis or reading was chosen after seeing the results it concerns, even when it was computed on confirmatory data. In Studies 1 and 2, uncertainty is reported as means over three seeds with 95\% confidence intervals from a paired bootstrap over trajectories; the intervals do not include seed variance. Study 3 reports per-seed outcomes and an exact binomial interval. Some exploratory cells rest on one or two seeds, and are marked.

\section{Study 1: does a JEPA organize its latent around fluid quantities?}\label{sec:4}

It does not, by the pre-registered criteria. The JEPA beats the untrained control on three probe targets but fails all three criteria (C, primary), and two controls that never see real physics read fluid quantities as well as or better than it does (C, descriptive).

\subsection{Models}\label{sec:4.1}

All six models share the encoder of Section~\ref{sec:3.2} and were trained with three seeds each.

\begin{table}[htbp]
\centering
\small
\begin{tabularx}{\linewidth}{l>{\raggedright\arraybackslash}Xl}
\toprule
Model & Training signal & Role \\
\midrule
JEPA & Predicts the target encoder's $8 \times 8$ feature map of frame $t + k$ from frames $(t - 1, t)$, $k \in \{1, 4\}$; EMA target; VICReg variance and covariance terms; no pixel loss & Main model \\
Pixel predictor & Decodes the velocity field of frame $t + k$; mean squared error & Comparison \\
Untrained & None (the Kaiming initialization trained models start from) & Control \\
Pooled JEPA & As the JEPA, but predicts the pooled 128-d vector & Ablation \\
Static VICReg & Invariance between two copies of one frame under random shifts and square symmetries; never sees time & Control \\
Non-physical JEPA & As the JEPA, trained on the same initial fields evolved by random damped, phase-rotating linear dynamics on each Fourier mode, with a hidden rate per trajectory & Control \\
\bottomrule
\end{tabularx}
\end{table}

The non-physical dynamics conserve no quadratic or Casimir quantity.

\subsection{Pre-registered criteria}\label{sec:4.2}

Three criteria compared the JEPA with the untrained control on held-out in-distribution trajectories, using the pooled latent.

\begin{itemize}
\item \textbf{S1 (structure).} Five probe targets: log enstrophy, log palinstrophy, log(1 + vortex count), the Okubo--Weiss fraction and log dissipation. A target counts if the JEPA's $R^2$ gain is at least 0.20 or halves the control's unexplained variance, with the bootstrap interval on the gain excluding zero. At least three targets must count, and one of them must be the Okubo--Weiss fraction. That clause was fixed before training so that a gain in amplitude-like quantities could not pass for learned geometry.
\item \textbf{S2 (hidden parameter).} Log $\nu$ probed from the latents of two consecutive frames: $R^2 \geq$ 0.6 and at least 0.3 above the control.
\item \textbf{S3 (prominence).} Enstrophy $|r|$ of the slowest directions (Section~\ref{sec:3.3}) at least 0.8, with the control below 0.8, a gap of at least 0.1, and the gap's interval excluding zero.
\end{itemize}

The hypothesis is supported if all three hold, partially supported if S1 holds and S2 or S3 fails, and not supported if S1 fails. A further rule stated that any S1 gain which disappears when the latent is averaged over all 64 sub-stride shifts of the input is attributed to shift tolerance rather than physical structure.

\subsection{Verdict}\label{sec:4.3}

\begin{table}[htbp]
\centering
\small
\caption{Study 1 verdict, in-distribution test set, pooled latent, mean of 3 seeds, 95\% paired trajectory-bootstrap intervals. The pre-registered verdict is \textbf{not supported} (C, primary).}
\label{tab:1}
\begin{tabularx}{\linewidth}{L{4.3cm}ccL{2.3cm}>{\raggedright\arraybackslash}X}
\toprule
Criterion & JEPA & Untrained & Needed & Outcome \\
\midrule
S1 log enstrophy & 0.930 & 0.727 & 0.864 & counts: $+0.204$ [$+0.175$, $+0.235$] \\
S1 log palinstrophy & 0.746 & 0.437 & 0.637 & counts: $+0.310$ [$+0.254$, $+0.366$] \\
S1 log(1 + vortex count) & 0.752 & 0.531 & 0.731 & counts: $+0.221$ [$+0.178$, $+0.275$] \\
S1 Okubo--Weiss fraction (mandatory) & 0.507 & 0.456 & 0.656 & does not count: $+0.051$ [$+0.002$, $+0.110$] \\
S1 log dissipation & 0.176 & 0.134 & 0.334 & does not count: $+0.042$ [$-0.005$, $+0.085$] \\
\textbf{S1} &  &  & 3 of 5, incl. Okubo--Weiss & \textbf{fails} (3 of 5, Okubo--Weiss not among them) \\
\textbf{S2} $\log\nu$, frame pair ($R^2$) & 0.457 & 0.481 & 0.781 & \textbf{fails} \\
\textbf{S3} enstrophy $|r|$ & 0.356 & 0.514 & 0.800 & fails; gap [$-0.31$, $+0.13$] \\
\bottomrule
\end{tabularx}
\end{table}

Under the shift-averaged readout the vortex-count gain falls to $+0.176$ and no longer counts, so by the pre-registered rule it is attributed to shift tolerance. Two attributable S1 gains remain, for enstrophy and palinstrophy. For energy, S3 cannot be passed at all: the untrained control's best slow direction already tracks it at $|r|$ 0.948, because energy is half the mean square of the input velocity.

\begin{figure}[htbp]
\centering
\includegraphics[width=\linewidth]{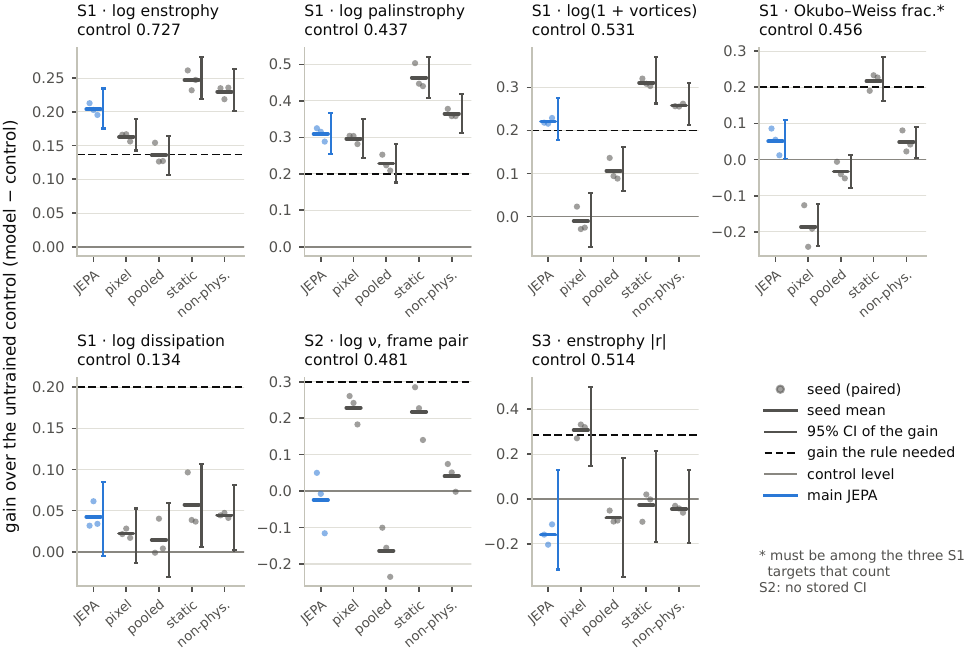}
\caption{Each model's gain over the untrained control on every pre-registered S1--S3 quantity (test set, pooled latent): seeds, seed mean and stored 95\% paired trajectory-bootstrap interval, against the gain each rule needed. The JEPA fails S1 (Okubo--Weiss), S2 and S3; static VICReg and the non-physical JEPA match or beat it (C).}
\label{fig:1}
\end{figure}

\subsection{The controls explain the JEPA's gains}\label{sec:4.4}

\begin{table}[htbp]
\centering
\small
\caption{Paired differences in probe $R^2$ (and S3 $|r|$), JEPA minus each comparison model; negative means the comparison model is better (C, descriptive).}
\label{tab:2}
\begin{tabularx}{\linewidth}{>{\raggedright\arraybackslash}Xccc}
\toprule
Quantity & JEPA $-$ non-physical JEPA & JEPA $-$ static VICReg & JEPA $-$ pooled JEPA \\
\midrule
log enstrophy & $-0.026$ [$-0.034$, $-0.020$] & $-0.043$ [$-0.056$, $-0.034$] & $+0.068$ [$+0.056$, $+0.082$] \\
log palinstrophy & $-0.055$ [$-0.075$, $-0.038$] & $-0.154$ [$-0.192$, $-0.124$] & $+0.081$ [$+0.062$, $+0.104$] \\
log(1 + vortex count) & $-0.037$ [$-0.052$, $-0.023$] & $-0.090$ [$-0.113$, $-0.068$] & $+0.115$ [$+0.092$, $+0.141$] \\
Okubo--Weiss fraction & $+0.003$ [$-0.042$, $+0.051$] & $-0.166$ [$-0.233$, $-0.109$] & $+0.084$ [$+0.042$, $+0.135$] \\
log dissipation & $-0.002$ [$-0.031$, $+0.029$] & $-0.015$ [$-0.072$, $+0.047$] & $+0.028$ [$-0.006$, $+0.061$] \\
$\log\nu$, frame-pair $R^2$ (no CI) & $-0.066$ & $-0.242$ & $+0.139$ \\
S3 enstrophy $|r|$ & $-0.114$ [$-0.21$, $+0.06$] & $-0.131$ [$-0.22$, $-0.02$] & $-0.076$ [$-0.11$, $+0.09$] \\
\bottomrule
\end{tabularx}
\end{table}

Predicting real Navier--Stokes dynamics added nothing measurable over predicting arbitrary smooth dynamics. The non-physical JEPA reads enstrophy, palinstrophy and vortex count significantly better than the JEPA, and is no worse on the other targets. The static VICReg model, which never sees time, is the strongest model on every S1 target and the only one that meets S1 against the control (4 of 5, including Okubo--Weiss). What the JEPA learned over random initialization is consistent with generic self-supervised feature quality, and the most useful ingredient appears to be the symmetry-invariance objective (or its augmentations) rather than prediction. Two caveats temper the comparison: the non-physical data change more slowly per frame, which makes that prediction task easier, and the static model sees augmentations the JEPA never sees. Predicting the spatial map rather than the pooled vector was the better JEPA design on every S1 target and on $\nu$ (significantly on four of five S1 targets), though the S3 point estimate favoured the pooled version; it was still not better than the controls.

\subsection{The pixel predictor and robustness}\label{sec:4.5}

The JEPA and the pixel predictor keep different information (C, descriptive). The JEPA keeps local structure: vortex count (0.752 against 0.521) and the Okubo--Weiss fraction (0.507 against 0.270). The pixel predictor keeps amplitude-like quantities: $\log\nu$ from a frame pair (0.710 against 0.457) and an enstrophy-tracking slow direction. Under the S3 rule the pixel predictor reaches 0.821 against 0.514 for the control, with gap interval [$+0.15$, $+0.50$] (PH: reported, unscored and not predicted). Training state does not explain the difference: the pixel predictor's validation loss was flat at 0.035 times its copy-the-frame baseline, while the JEPA's was at 0.60 times its baseline and still falling slowly at epoch 40.

The verdict does not hinge on the exact rule (PH). The literal wording of the original brief counts 1 target; the pre-registered rule counts 3 without Okubo--Weiss; the shift-averaged readout counts 2. Dropping the Okubo--Weiss clause would give 3 of 5 and a ``partially supported'' label, but the same rule would then also pass the non-physical JEPA (3 of 5) and static VICReg (4 of 5). No model breaks down on the held-out Reynolds band: errors there are 0.71--1.46 times those on neighbouring in-distribution bands, for every model and target (C, descriptive).

\section{Study 2: what the training objective does}\label{sec:5}

Models trained to render velocity or vorticity without a whitening regularizer make enstrophy prominent, and predicting dynamics is not required; adding the regularizer reduces that prominence in every comparison, though not always by the pre-registered margin, while raising the autoencoders' decodability. On sealed data with fresh seeds, the rendering hypothesis holds by a wide margin and the regularization hypothesis holds in direction but not at the pre-registered size for one render target, so the verdict is partially supported (C).

These hypotheses came from an exploratory line on fresh data, summarized in Figure~\ref{fig:2} and described in Appendix~\ref{sec:5.1}, in which we call the pixel predictor the ``imaginer'' and a JEPA trained at the same horizons the ``summarizer''.

\begin{figure}[htbp]
\centering
\includegraphics[width=\linewidth]{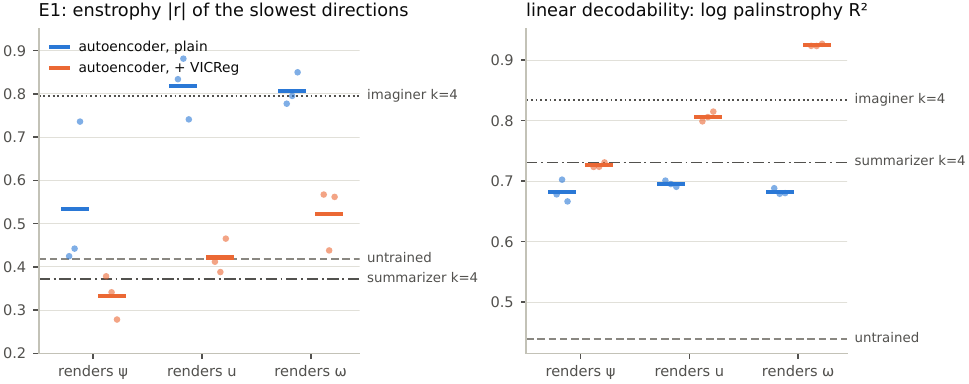}
\caption{Plain autoencoders that render the present match the imaginer's enstrophy prominence; adding VICReg drops it to control level while raising linear decodability (E). The E4 panel is Figure~\ref{fig:B1} (Appendix~\ref{app:B}).}
\label{fig:2}
\end{figure}

\subsection{Confirmatory hypotheses}\label{sec:5.2}

Four hypotheses were pre-registered and evaluated once on a sealed set of 1,005 trajectories (378 inviscid). The primary hypotheses used freshly trained seeds that had never been evaluated.

\begin{itemize}
\item \textbf{H1 (primary): rendering makes enstrophy prominent.} For a plain autoencoder rendering velocity and one rendering vorticity: E1 $\geq$ 0.70, at least 0.25 above the untrained control, interval excluding zero. Both must pass.
\item \textbf{H2 (primary): the regularizer reduces it.} For both render targets, plain minus VICReg E1 $\geq$ 0.20, interval excluding zero.
\item \textbf{H3 (secondary): the render target sets the mix.} For each seed, take the enstrophy-tracking direction, the one of the three slowest directions with the largest $|r|$ with $\log Z_0$. Regress its trajectory-mean projection by least squares on an intercept, $\log E_0$ and $\log(Z_0/E_0)$, unstandardized. E4~$= a/b$ is the coefficient on $\log E_0$ divided by the coefficient on $\log(Z_0/E_0)$, averaged over seeds. Because $\log Z_0 = \log E_0 + \log(Z_0/E_0)$, E4~$= 1$ means the direction tracks $\log Z_0$ alone and E4~$> 1$ means it leans toward energy. H3 holds if the vorticity autoencoder's E4 is closer to 1 on a log scale than the streamfunction autoencoder's, and the streamfunction autoencoder's E4 exceeds 2 (exploration seeds 0--2).
\item \textbf{H4 (secondary): the regularizer explains the imaginer--summarizer gap.} Imaginer ($k = 4$) minus imaginer with VICReg $\geq$ 0.20, interval excluding zero.
\end{itemize}

\subsection{Results}\label{sec:5.3}

\begin{table}[htbp]
\centering
\small
\caption{Study 2 confirmation on the sealed set; untrained control E1 0.380 [0.30, 0.46]. The pre-registered verdict is \textbf{partially supported}: H1 holds and H2 fails (C, primary; H3 and H4 secondary).}
\label{tab:3}
\begin{tabularx}{\linewidth}{>{\raggedright\arraybackslash}X>{\raggedright\arraybackslash}X>{\raggedright\arraybackslash}Xl}
\toprule
Hypothesis & Rule & Result & Outcome \\
\midrule
H1, velocity autoencoder & E1 $\geq$ 0.70; $\geq$ control + 0.25; CI > 0 & E1 0.860; $+0.480$ [$+0.390$, $+0.565$] & pass \\
H1, vorticity autoencoder & same & E1 0.799; $+0.418$ [$+0.360$, $+0.486$] & pass \\
\textbf{H1} & both pass &  & \textbf{holds} \\
H2, velocity: plain $-$ VICReg & $\geq$ 0.20; CI > 0 & $+0.338$ [$+0.289$, $+0.390$] & pass \\
H2, vorticity: plain $-$ VICReg & same & $+0.149$ [$+0.105$, $+0.193$] & \textbf{fail} \\
\textbf{H2} & both pass &  & \textbf{fails} \\
H3 (secondary), E4: vorticity vs streamfunction & $\omega$ closer to 1; $\psi > 2$ & 1.00 vs 3.13 & pass \\
H4 (secondary), imaginer $-$ imaginer + VICReg & $\geq$ 0.20; CI > 0 & $+0.315$ [$+0.272$, $+0.358$] & pass \\
\bottomrule
\end{tabularx}
\end{table}

Per seed, E1 was 0.894, 0.856 and 0.830 for the velocity autoencoder and 0.709, 0.837 and 0.850 for the vorticity autoencoder; with VICReg, 0.568, 0.501 and 0.497 (velocity) and 0.621, 0.649 and 0.679 (vorticity). The vorticity shortfall comes from one weaker plain seed (0.709) and VICReg seeds that kept more prominence than the exploration seeds had. The exploration seeds would have passed H2 on the same sealed data ($+0.262$), which is exactly the training-run variability that fresh seeds were added to expose (C, descriptive).

\begin{figure}[htbp]
\centering
\includegraphics[width=\linewidth]{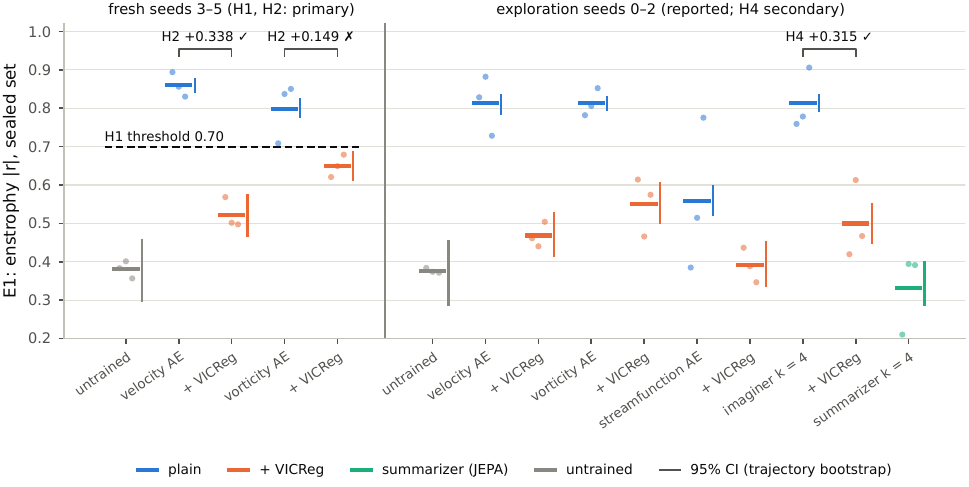}
\caption{Enstrophy prominence E1 on the sealed Study 2 set (seeds, mean, 95\% trajectory-bootstrap interval). Fresh seeds 3--5 decide H1 and H2; exploration seeds 0--2 are reported and carry H4 (C).}
\label{fig:3}
\end{figure}

\textbf{Prominence and decodability dissociate.} VICReg lowered E1 in all six plain--VICReg pairs on the sealed set (the five with computed intervals all exclude zero), while every VICReg autoencoder decoded log enstrophy and log palinstrophy better than its plain counterpart (the imaginer with VICReg did not: 0.937 and 0.782 against 0.941 and 0.830). For the velocity autoencoder, log enstrophy $R^2$ rose from 0.862 to 0.933 and log palinstrophy from 0.700 to 0.799; for the vorticity autoencoder, from 0.806 to 0.955 and from 0.694 to 0.912 (C, descriptive). One reading consistent with this is that the regularizer keeps the information but spreads it across the latent, so that it no longer dominates the slowest directions.

\textbf{A confound we cannot remove.} VICReg also made rendering about 25 times worse for the autoencoders and about 20 times worse for the pixel predictor. Part of each drop in prominence may therefore come from worse rendering rather than from latent geometry, and this design cannot separate the two.

\subsection{What this says about Study 1}\label{sec:5.4}

The imaginer--summarizer gap at $k = 4$ replicated on the sealed set ($+0.483$ [$+0.42$, $+0.52$]; C, descriptive), and adding VICReg to the imaginer removed $+0.315$ of it, about two-thirds (H4; C, secondary). Study 1's JEPA result is therefore \emph{consistent with} its regularizer. It is not shown to be \emph{caused} by it: a JEPA without VICReg collapses, so the summarizer could not be tested unregularized. Predicting dynamics was never required for the effect, since autoencoders rendering the present show it.

\section{Study 3: building steerable dials}\label{sec:6}

A decoder can be given label-free dials for loudness and spectral shape; in our build, dials appeared only once we controlled which pathway carried each property to the output. On sealed data, the pre-registered rules find a loudness dial in all 8 trained models and a shape dial in 3 of 5 fresh seeds of the final design, against a bar of 4, while neither control shows a shape dial. The verdict is partially supported (C). Sections~\ref{sec:6.1}--\ref{sec:6.4} describe the exploratory build that led to the design (E); Section~\ref{sec:6.5} is the confirmation (C); Appendix~\ref{sec:6.6} is a post hoc analysis of its failures (PH).

\subsection{Readable is not steerable}\label{sec:6.1}

We first measured 21 existing models from the Study 2 exploration (autoencoders, VICReg autoencoders and imaginers, seeds 0--2), taking the first four principal components of each pooled latent. None of the 84 candidate directions moved the decoded field by even 0.1 log-units per unit $\alpha$, the threshold at the time; the largest effect was 0.086. A supervised direction fitted to predict spectral shape $s$ did predict it almost perfectly ($|r|$ 0.97--0.99; an untrained encoder reads it nearly as well, 0.96), yet pushing the latent along it barely changed the rendered field (E). The information was readable and not steerable, which set the goal for the build: make a direction that the decoder actually uses.

\begin{figure}[htbp]
\centering
\includegraphics[width=\linewidth]{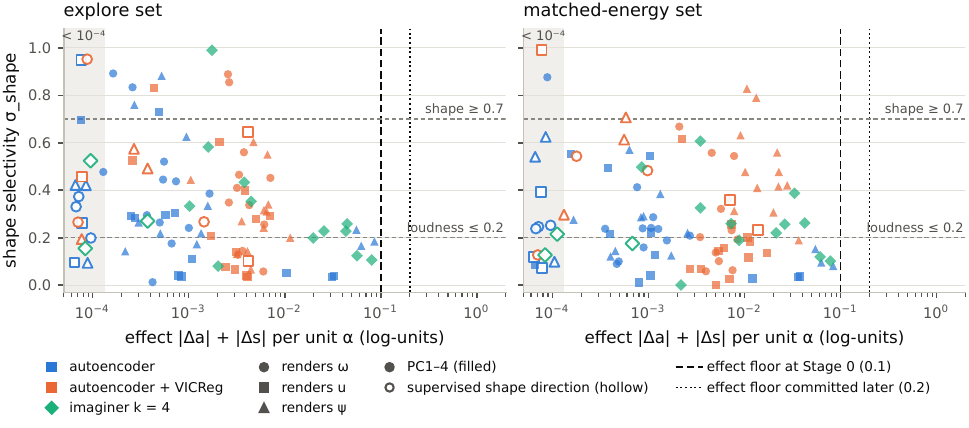}
\caption{Of 84 candidate latent directions in 21 existing models, none moves the decoded field by the dial threshold; even the supervised shape direction (hollow markers), which predicts $s$ almost perfectly, barely moves it. Dashed line: the 0.1 floor used at the time; dotted: the 0.2 floor committed afterwards. Effects below $10^{-4}$ are drawn in the shaded strip (E).}
\label{fig:4}
\end{figure}

\subsection{Design: route each property through one pathway}\label{sec:6.2}

The final design splits the encoder into two branches whose outputs reach the decoder by different routes.

\begin{itemize}
\item \textbf{Dial code.} A separate dial encoder, of the same design and reading the raw input, produces an 8-dimensional pooled vector. All eight dials modulate every decoder stage (the input stage and each of three upsampling stages) through zero-initialized feature-wise gain and bias (FiLM; \citealp{perez2018film}). Because each decoder stage builds a different range of spatial scales, per-stage gains were intended to act like a spectral equalizer, so that a change in spectral shape is something the code can express.
\item \textbf{Scale-free content.} The content branch sees the input normalized to unit root-mean-square amplitude, so overall loudness can reach the decoder only through the dial code. That loudness reaches the decoder only through the dials is therefore a design property; the tested claim is that label-free rules find both dials and that both pass every realism criterion.
\item \textbf{Shape noise on the content.} Before the content branch sees it, the RMS-normalized input's spectrum is distorted by a random low-order gain on shells 1--21, with log-amplitude $\tau(\epsilon_1\sigma_1\phi_1 + \epsilon_2\sigma_2\phi_2)/2$. Here $\phi_1$ is the centred $\log k$ (tilt), $\phi_2$ the centred square of $\phi_1$ (curvature), $\sigma_1$ and $\sigma_2$ the training data's own spread of those coefficients, $\epsilon_1$ and $\epsilon_2$ standard normal, and $\tau$ the dose. The mean and every shell above 21 (including the corners of the square band) are then zeroed, and the input is RMS-normalized again. Draws are fresh at every training step and fixed per frame at evaluation. The low-order form targets the coarse spectral shape the content would otherwise use.
\item \textbf{Decoder and loss.} The decoder upsamples by nearest-neighbour resizing followed by two $3 \times 3$ convolutions per stage (1.45 M parameters, 79\% of the earlier sub-pixel decoder). The loss is the vorticity rendering error (weight 14.7) plus VICReg variance (25) and covariance (1) terms on the pooled 120-dimensional content latent, so, in line with Study 2, the content code is regularized with VICReg and the dial code is not. Training uses AdamW (learning rate $10^{-3}$, weight decay 0.05, batch 256) for 30 epochs with 5\% warmup, cosine decay and gradient clipping at 1.0.
\end{itemize}

Appendix~\ref{app:C} describes how this design was reached, step by step.

\subsection{Identifying dials by their effect}\label{sec:6.3}

A dial is only useful to a downstream user if it can be found without labels, so every identification rule had to be label-free. The obvious rule, principal components of the dial code, fails for a simple reason: in this data loudness and shape are strongly correlated ($r = -0.82$), so the directions of largest variance are mixtures of the two. PCA ranks directions by how the data vary, not by what the directions do.

\textbf{Loudness by symmetry.} We find the loudness dial as the direction of the dial code whose steered output is closest to a pure rescaling of the unsteered output. The objective is $J = \sum\|y_\alpha - c\, y_0\|^2 / \sum\|y_\alpha - y_0\|^2$, the fraction of the steered change not explained by the best scalar rescaling $c$ of the unsteered output $y_0$ (fitted per frame and $\alpha$), minimized over unit directions. Normalizing by the size of the change matters: the raw residual is minimized by a direction that changes nothing. This uses no physics labels, only the generic definition of ``louder''. In the whitened-content model it left 3.5--3.7\% of the steered change unexplained, against 31--59\% for the first principal component. The direction was pure loudness ($\sigma_{\mathrm{shape}}$ 0.02--0.08) with no artefact frames in all three seeds, and a valid dial in two (the third missed the cycle gain, 0.48). The direction it finds lies only partly in the plane of the first two principal components (65--79\% of it).

\textbf{Loudness-null PCA.} The rule we carried into confirmation estimates $g$, the average gradient of output loudness $a$ with respect to the dial code (finite differences of $\pm 0.5 z$-units per coordinate, averaged over 256 exploration frames), restricts the $z$-scored dial code to the subspace orthogonal to $g$, where steering leaves output loudness unchanged to first order, and takes the leading principal component of the data in that subspace. It uses loudness, a generic notion, and finds shape from the data. During exploration we kept the second principal component (PC2) as the frozen primary shape rule and reported loudness-null PCA alongside it; loudness-null PCA was therefore chosen after seeing results, which is why only a fresh confirmation could test it.

Two other shape rules failed: the leading principal component after projecting loudness out of the latent, and the directions of largest raw output sensitivity (Appendix~\ref{app:C}).

\subsection{Dose, decoder and cost}\label{sec:6.4}

\begin{table}[htbp]
\centering
\small
\caption{Shape-noise dose (exploration data, 3 seeds per row). ``Content shape $R^2$'' is how well $s$ can be read from the content branch's input; ``shape knob'' is the loudness-null direction (E).}
\label{tab:4}
\begin{tabularx}{\linewidth}{cccccc}
\toprule
$\tau$ & Content shape $R^2$ & Decoder & Shape-knob $\sigma_{\mathrm{shape}}$ (median) & Both dials, all criteria & $\omega$ error \\
\midrule
0 & 0.97 & sub-pixel & 0.07 & 0 of 3 & 0.015 \\
0.2 & 0.61 & resize & 0.88 & 3 of 3 & 0.022 \\
0.5 & 0.29 & sub-pixel & 0.93 & 0 of 3 (artefacts) & 0.041 \\
0.5 & 0.29 & resize & 0.96 & 3 of 3 & 0.039 \\
1 & 0.18 & sub-pixel & 0.88 & 0 of 3 & 0.125 \\
2 & 0.07 & sub-pixel & 0.93 & 0 of 3 & 0.374 \\
\bottomrule
\end{tabularx}
\end{table}

\textbf{A dial forms when the content's copy is degraded, not only when it is removed.} With no noise, the content reads shape at $R^2$ 0.97 and the loudness-null direction is not a shape knob ($\sigma_{\mathrm{shape}}$ 0.07). At $\tau = 0.2$ the content still reads shape at $R^2$ 0.61, yet the loudness-null direction is already almost pure shape ($\sigma_{\mathrm{shape}}$ 0.88). One interpretation is that the decoder takes shape from the dial code once that route is the more reliable one (E). We do not claim a threshold or jump: the curve has one intermediate point and spans two decoders.

\textbf{Decoder upsampling decides whether strong dials stay clean.} Strong dials built on the sub-pixel decoder repeatedly produced grid artefacts at the extremes of the steering range, and at $\tau = 0.5$ the artefact criterion removed every model that otherwise had both dials. Earlier, the scale-free model's loudness dials had turned 5--14\% of frames into checkerboards at $\alpha = +2$; at $\tau = 0.5$, shape directions broke up to 9.4\% of frames and loudness directions up to 4.3\%. These dials looked clean in median-based criteria and failed only in the tails. In exploration, replacing sub-pixel upsampling with resize-convolution removed the artefacts completely (0.0\% of frames, every direction, every seed at $\tau = 0.5$), and the shape knob survived the change ($\sigma_{\mathrm{shape}}$ 0.86--1.00 against 0.92--1.00). The new decoder differs in depth and parameter count as well as in upsampling, so we credit the decoder as a whole. On sealed data one resize-decoder direction still exceeded the criterion (seed 102's shape direction, Section~\ref{sec:6.5}), and untrained resize-decoder networks produced artefacts in every frame.

Appendix~\ref{app:C} shows the artefacts (Figure~\ref{fig:5}) and the dose curve (Figure~\ref{fig:6}).

\textbf{The cost is rendering.} The final exploration model ($\tau = 0.2$, resize decoder) renders vorticity with normalized error 0.022: 1.5 times its $\tau = 0$ version, 2.1 times the VICReg autoencoder and 19 times the plain autoencoder. In exploration it gave both dials in 3 of 3 seeds under loudness-null PCA and 2 of 3 under PC2, and a dose rule committed before training selected it over $\tau = 0.5$ for confirmation.

\subsection{Confirmation on sealed data}\label{sec:6.5}

The pre-registration fixed the model ($\tau = 0.2$, resize decoder), five fresh main seeds, two controls and the rules. The controls were a $\tau = 0$ model with the same decoder (three fresh seeds), allowed by rule to show a shape dial in at most 1 of 3 seeds, and an untrained network with randomly initialized FiLM layers (three seeds), allowed none. The loudness rule was the symmetry rule and the primary shape rule was loudness-null PCA; PC2 was secondary, with its exploration record (1 of 3, then 2 of 3) stated. Every decoded field had its spatial mean removed before measurement, because steering occasionally added a mean that a periodic flow cannot have; the identification rules themselves used the raw decoder output. Two sealed sets were opened once each: a general set (1,007 trajectories, 378 inviscid) and a new matched-energy set (300 inviscid). The design passed if both dials were valid under every criterion on both sets in at least 4 of 5 main seeds, was partial at 3, and failed at 2 or fewer. Rendering was to be reported, not gated.

\begin{table}[htbp]
\centering
\footnotesize
\caption{Study 3 confirmation, per fresh seed (C, primary). The exact 95\% interval for a rate observed as 3 of 5 is 15--95\%.}
\label{tab:5}
\begin{tabularx}{\linewidth}{C{0.85cm}C{1.25cm}C{1.0cm}C{0.9cm}C{2.0cm}C{2.0cm}>{\raggedright\arraybackslash}X}
\toprule
Seed & Loudness dial & Shape dial & Both & $\sigma_{\mathrm{shape}}$ loudness / shape, general & $\sigma_{\mathrm{shape}}$ loudness / shape, matched energy & What failed \\
\midrule
100 & yes & yes & \textbf{yes} & 0.00 / 0.94 & 0.03 / 0.94 & --- \\
101 & yes & yes & \textbf{yes} & 0.00 / 0.98 & 0.04 / 0.78 & --- \\
102 & yes & no & no & 0.00 / 0.41 & 0.05 / 0.46 & shape, both sets: effect, monotonicity, cycle correlation and gain, grid artefacts (3.5\% / 4.7\% of frames); envelope on matched energy \\
103 & yes & yes & \textbf{yes} & 0.00 / 1.00 & 0.05 / 0.81 & --- \\
104 & yes & no & no & 0.00 / 0.96 & 0.03 / 0.96 & shape, matched energy only: envelope $-0.09$ (tolerance $-0.05$) \\
\textbf{Count} & 5 of 5 & 3 of 5 & \textbf{3 of 5} &  &  & pass $\geq$ 4 $\rightarrow$ \textbf{partially supported} \\
\bottomrule
\end{tabularx}
\end{table}

The pre-registered verdict is partially supported, with its wording fixed in advance: the design produces both dials ``in most, but not reliably in all, training runs; it would need a more robust identification rule or training before use.'' Where valid, the dials were strong and clean. Loudness dials had effect 0.57--0.65 per unit $\alpha$ with $\sigma_{\mathrm{shape}}$ at most 0.05; the three valid shape dials had effect 0.33--0.46 with $\sigma_{\mathrm{shape}}$ 0.78--1.00; every valid dial had monotonicity 1.00, cycle correlation 0.95--1.00 and cycle gain 0.60--0.97.

Both controls stayed within their rules. The $\tau = 0$ models had a loudness dial in 3 of 3 seeds and a shape dial in none: their loudness-null directions moved the field by only 0.02--0.10 per unit $\alpha$. The untrained networks had no dials of either kind, and their steered fields carried grid artefacts in every frame. The shape dial therefore appeared only when the content pathway's copy of shape was degraded, and never in an untrained network (C, primary).

With PC2 as the shape rule, both dials were valid in 3 of 5 seeds, but in seeds 100, 102 and 104 (C, secondary). The main model rendered vorticity with error 0.021--0.023, about 2.2 times the VICReg autoencoder and 1.5 times its $\tau = 0$ version. In one passing seed (103), the zero-mean step removed a sizeable spatial-mean offset at $\alpha = -2$ in 23\% of frames of the general set; whether that seed would pass without the step was not computed on sealed data (C, descriptive).

\begin{figure}[htbp]
\centering
\includegraphics[width=\linewidth]{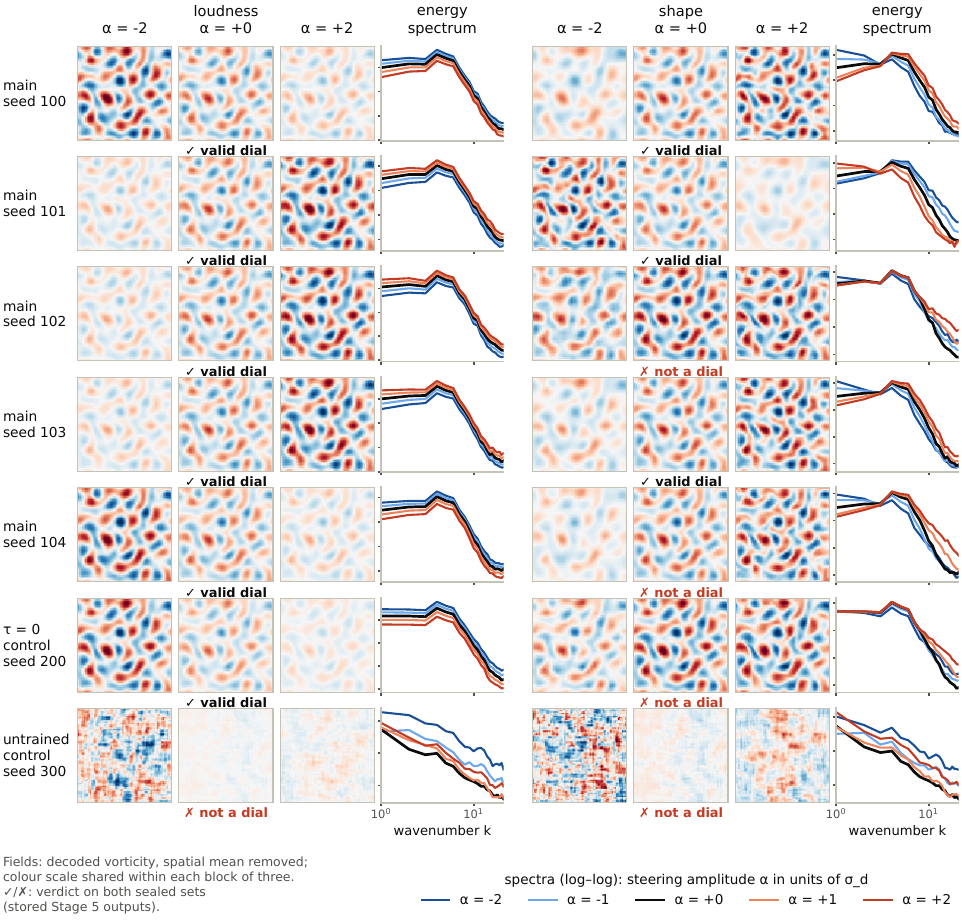}
\caption{Both dials of the five fresh main models and one seed of each control, decoded on one exploration frame (the sealed sets are spent and are not reopened for plotting); \ding{51}/\ding{55} give each direction's verdict on both sealed sets. Loudness scales the field; shape moves energy between scales; seed 102's shape direction fails. All 11 models: Figure~\ref{fig:D1} (Appendix~\ref{app:D}) (C).}
\label{fig:7}
\end{figure}

\section{Discussion}\label{sec:7}

Across the three studies, the properties usually read as evidence of physical understanding came apart, and each depended in our experiments more on design choices than on what the model was asked to predict.

\subsection{Decodable, prominent and steerable are different properties}\label{sec:7.1}

The separation appeared at every level. The JEPA decoded fluid quantities well, but no better than models trained on fake physics or no time at all (C, descriptive). VICReg raised the autoencoders' decodability while lowering prominence (C, descriptive). A supervised direction read spectral shape almost perfectly and moved nothing when the latent was pushed along it (E). In the whitened-content model (Stage 2b), the supervised shape direction also failed the round trip (cycle gain 0.20--0.31, below 0.5; E). Claims that a model has learned a physical quantity should therefore say which of the three properties they test, and a claim that a model \emph{uses} a quantity needs an intervention on the latent, not a probe.

\subsection{Objective geometry, not dynamics}\label{sec:7.2}

In this setting, enstrophy became prominent when a model had to render velocity or vorticity and its latent was not whitened; rendering the streamfunction gave a weaker, energy-leaning direction (E1 0.558; C, descriptive). Predicting dynamics was not required: autoencoders rendering the present frame showed the effect as strongly as models rendering the future (C, descriptive). The JEPA's weak showing in Study 1 is consistent with its regularizer, which, added to the pixel predictor, removed about two-thirds of its advantage over the summarizer at $k = 4$ (C, secondary). For evaluating world models, a slow or principal direction that tracks a conserved quantity is weak evidence of learned physics. Such claims should at least be compared against a present-frame autoencoder and a swap of the regularizer.

\subsection{Routing as an inductive bias for control}\label{sec:7.3}

Unsupervised disentanglement needs inductive biases (\citealp{locatello2019challenging}). The bias that worked here is architectural: a dial formed for a property when the dial code was the more reliable route for that property to reach the output (E, interpretation). Loudness had a dial in all eight trained models once the content pathway could not see scale (C). Shape had one in most seeds once the content pathway's copy of it was degraded, even though that copy was still partly readable (C; $R^2$ 0.61 from exploration, E).

This is an instrument panel designed in advance, not one that grows by itself: dials appear only for properties someone chose to route. What the designer supplies, however, is a transformation that varies the property (a rescaling, a random spectral tilt), not a formula for it. The loudness dial was also \emph{identified} by its transformation. Whether a single transformation can both remove a property from the content path and identify its dial, for a property with no closed-form handle, is the open question this work points to. Our one attempt to use the transformation for identification (Appendix~\ref{sec:6.6}) found clean but weak directions, so it is not yet a working rule.

\subsection{Practical lessons}\label{sec:7.4}

These are mostly observations from exploration and post hoc analysis, not confirmed claims; the last is a confirmatory descriptive result.

\begin{itemize}
\item \textbf{Find dials by what they do.} For loudness, a rule based on the output (rescaling symmetry) beat rules based on data variance. For shape, the output-constrained rule beat the variance rule in exploration but tied with it on sealed data (3 of 5 each). Orthogonality in the latent did not imply independence in the output.
\item \textbf{Look at the tails.} Dials that passed median-based criteria broke 5--14\% of frames at the extremes of the steering range. A criterion for the tails had to be added, and it overturned earlier positives.
\item \textbf{Check the decoder's upsampling.} Sub-pixel upsampling repeatedly produced checkerboard artefacts under strong steering; resize-convolution removed them in exploration, though one direction on sealed data still exceeded the criterion.
\item \textbf{Use fresh seeds for confirmation.} The exploration seeds would have passed Study 2's H2 on the same sealed data; fresh seeds did not (C, descriptive).
\end{itemize}

\subsection{Limitations}\label{sec:7.5}

\begin{itemize}
\item \textbf{Two dimensions only.} Two-dimensional flow has no vortex stretching, its energy cascade runs to large scales (\citealp{kraichnan1967}), and enstrophy is an inviscid invariant only in 2D. Neither the positive nor the negative results are claimed to transfer to three-dimensional flow, other simulators or initial-condition distributions, or larger models.
\item \textbf{One small architecture.} All encoders are one small convolutional family without normalization layers (Study 3's content branch normalizes its input per sample by design), at $64 \times 64$ resolution, with three to five seeds per confirmatory configuration and as few as one or two in some exploratory cells. Intervals resample trajectories, not seeds.
\item \textbf{Method-relative prominence.} E1 is defined by one unsupervised pipeline (z-scoring, 32 principal components, a shrinkage within/between eigenproblem). Other pipelines could rank directions differently.
\item \textbf{A confound in Study 2.} VICReg also degraded rendering by roughly 20--25 times, which this design cannot separate from its effect on latent geometry.
\item \textbf{The dial design's limits.} The route by which loudness reaches the decoder exists by construction; finding a valid loudness dial without labels was the test. The shape dial needs a known transformation, costs rendering quality (2.2 times the VICReg autoencoder), and its identification rule was chosen during exploration and worked in 3 of 5 confirmation seeds. The dose curve has one intermediate point and spans two decoders.
\item \textbf{Adaptive criteria.} Some criteria were changed or calibrated on exploration data (the effect floor, the artefact criterion). Each change was committed before any confirmatory use (Appendix~\ref{sec:8.2}), but the criteria are not independent of the exploration.
\end{itemize}

\subsection{Future work}\label{sec:7.6}

Three next steps follow directly. First, a confirmation of the dial design with a single identification rule fixed in advance and at least ten fresh seeds, since 3 of 5 leaves the success rate anywhere between 15\% and 95\%, and even 8 of 10 would leave it between 44\% and 97\%. Second, a test of the transformation-defined recipe on a property without a closed-form handle, ideally as a registered report reviewed before it is run. Third, a coarse three-dimensional version, where vortex stretching makes the dynamics, and the quantities worth controlling, qualitatively different.

\section{Process, reproducibility and AI assistance}\label{sec:8}

The whole program ran over about two weeks under one protocol, and every departure from it is recorded. Appendix~\ref{app:F} gives the record and lists the deviations; the released timeline gives each one's date and commit.

\subsection{Release}\label{sec:8.4}

The repository contains the simulator and its tests, all training and evaluation code, every configuration, the pre-registrations, the prediction entries with their probabilities, the run log, all reports, and the checkpoint hashes. Generated data can be regenerated from the configurations and are checksummed. \ReleaseStatement

\subsection{AI assistance}\label{sec:8.5}

The author conceived the research question, wrote the brief for each line of work, and approved every hypothesis, decision rule, control, budget and sealed-set opening. Several designs and rule definitions were proposed by AI assistants and accepted, modified or rejected by the author; verdict texts were fixed in advance and applied by the frozen scripts. An AI assistant (Anthropic's Claude), used in conversation, proposed and critiqued experimental designs and drafted brief text. An AI coding agent (Claude Code) drafted the pre-registrations in the repository for the author's approval before each commit, wrote and ran the simulator, training and evaluation code, kept the run log, wrote the probabilities in the prediction entries (except those marked as the author's), and drafted the stage reports. This manuscript was drafted with AI assistance from the committed reports, then audited claim by claim against the committed outputs by the AI coding agent, which had not written the manuscript, and every reference was checked against its source. The author takes full responsibility for the content, consistent with TMLR's policy that language models may be used as general-purpose assistive tools.

\label{endofmaintext}
\bibliography{refs}

\begin{thebibliography}{27}
\providecommand{\natexlab}[1]{#1}
\providecommand{\url}[1]{\texttt{#1}}
\expandafter\ifx\csname urlstyle\endcsname\relax
  \providecommand{\doi}[1]{doi: #1}\else
  \providecommand{\doi}{doi: \begingroup \urlstyle{rm}\Url}\fi

\bibitem[Aitken et~al.(2017)Aitken, Ledig, Theis, Caballero, Wang, and
  Shi]{aitken2017checkerboard}
Andrew Aitken, Christian Ledig, Lucas Theis, Jose Caballero, Zehan Wang, and
  Wenzhe Shi.
\newblock Checkerboard artifact free sub-pixel convolution: A note on sub-pixel
  convolution, resize convolution and convolution resize.
\newblock arXiv:1707.02937, 2017.

\bibitem[Albanie et~al.(2022)Albanie, Henriques, Bertinetto,
  Hern{\'a}ndez-Garc{\'i}a, Doughty, and Varol]{albanie2022prereg}
Samuel Albanie, Jo{\~a}o~F. Henriques, Luca Bertinetto, Alex
  Hern{\'a}ndez-Garc{\'i}a, Hazel Doughty, and G{\"u}l Varol (eds.).
\newblock \emph{NeurIPS 2021 Workshop on Pre-registration in Machine Learning},
  volume 181 of \emph{Proceedings of Machine Learning Research}, 2022. PMLR.
\newblock URL \url{https://proceedings.mlr.press/v181/}.

\bibitem[Alet et~al.(2021)Alet, Doblar, Zhou, Tenenbaum, Kawaguchi, and
  Finn]{alet2021noether}
Ferran Alet, Dylan Doblar, Allan Zhou, Josh Tenenbaum, Kenji Kawaguchi, and
  Chelsea Finn.
\newblock {Noether Networks}: Meta-learning useful conserved quantities.
\newblock In \emph{Advances in Neural Information Processing Systems},
  volume~34, pp.\  16384--16397, 2021.

\bibitem[Assran et~al.(2023)Assran, Duval, Misra, Bojanowski, Vincent, Rabbat,
  LeCun, and Ballas]{assran2023ijepa}
Mahmoud Assran, Quentin Duval, Ishan Misra, Piotr Bojanowski, Pascal Vincent,
  Michael Rabbat, Yann LeCun, and Nicolas Ballas.
\newblock Self-supervised learning from images with a joint-embedding
  predictive architecture.
\newblock In \emph{Proceedings of the IEEE/CVF Conference on Computer Vision
  and Pattern Recognition (CVPR)}, pp.\  15619--15629, 2023.
\newblock \doi{10.1109/CVPR52729.2023.01499}.

\bibitem[Bardes et~al.(2022)Bardes, Ponce, and LeCun]{bardes2022vicreg}
Adrien Bardes, Jean Ponce, and Yann LeCun.
\newblock {VICReg}: Variance-invariance-covariance regularization for
  self-supervised learning.
\newblock In \emph{International Conference on Learning Representations}, 2022.

\bibitem[Bardes et~al.(2024)Bardes, Garrido, Ponce, Chen, Rabbat, LeCun,
  Assran, and Ballas]{bardes2024vjepa}
Adrien Bardes, Quentin Garrido, Jean Ponce, Xinlei Chen, Michael Rabbat, Yann
  LeCun, Mido Assran, and Nicolas Ballas.
\newblock Revisiting feature prediction for learning visual representations
  from video.
\newblock \emph{Transactions on Machine Learning Research}, 2024.
\newblock URL \url{https://openreview.net/forum?id=QaCCuDfBk2}.

\bibitem[Belinkov(2022)]{belinkov2022probing}
Yonatan Belinkov.
\newblock Probing classifiers: Promises, shortcomings, and advances.
\newblock \emph{Computational Linguistics}, 48\penalty0 (1):\penalty0 207--219,
  2022.
\newblock \doi{10.1162/coli_a_00422}.

\bibitem[Bertinetto et~al.(2021)Bertinetto, Henriques, Albanie, Paganini, and
  Varol]{bertinetto2021prereg}
Luca Bertinetto, Jo{\~a}o~F. Henriques, Samuel Albanie, Michela Paganini, and
  G{\"u}l Varol (eds.).
\newblock \emph{NeurIPS 2020 Workshop on Pre-registration in Machine Learning},
  volume 148 of \emph{Proceedings of Machine Learning Research}, 2021. PMLR.
\newblock URL \url{https://proceedings.mlr.press/v148/}.

\bibitem[Elazar et~al.(2021)Elazar, Ravfogel, Jacovi, and
  Goldberg]{elazar2021amnesic}
Yanai Elazar, Shauli Ravfogel, Alon Jacovi, and Yoav Goldberg.
\newblock Amnesic probing: Behavioral explanation with amnesic counterfactuals.
\newblock \emph{Transactions of the Association for Computational Linguistics},
  9:\penalty0 160--175, 2021.
\newblock \doi{10.1162/tacl_a_00359}.

\bibitem[Greydanus et~al.(2019)Greydanus, Dzamba, and
  Yosinski]{greydanus2019hnn}
Samuel Greydanus, Misko Dzamba, and Jason Yosinski.
\newblock Hamiltonian neural networks.
\newblock In \emph{Advances in Neural Information Processing Systems},
  volume~32, 2019.

\bibitem[Hewitt \& Liang(2019)Hewitt and Liang]{hewitt2019control}
John Hewitt and Percy Liang.
\newblock Designing and interpreting probes with control tasks.
\newblock In \emph{Proceedings of the 2019 Conference on Empirical Methods in
  Natural Language Processing and the 9th International Joint Conference on
  Natural Language Processing (EMNLP-IJCNLP)}, pp.\  2733--2743, 2019.
\newblock \doi{10.18653/v1/D19-1275}.

\bibitem[Higgins et~al.(2017)Higgins, Matthey, Pal, Burgess, Glorot, Botvinick,
  Mohamed, and Lerchner]{higgins2017betavae}
Irina Higgins, Loic Matthey, Arka Pal, Christopher Burgess, Xavier Glorot,
  Matthew Botvinick, Shakir Mohamed, and Alexander Lerchner.
\newblock beta-{VAE}: Learning basic visual concepts with a constrained
  variational framework.
\newblock In \emph{International Conference on Learning Representations}, 2017.

\bibitem[Kraichnan(1967)]{kraichnan1967}
Robert~H. Kraichnan.
\newblock Inertial ranges in two-dimensional turbulence.
\newblock \emph{The Physics of Fluids}, 10\penalty0 (7):\penalty0 1417--1423,
  1967.
\newblock \doi{10.1063/1.1762301}.

\bibitem[LeCun(2022)]{lecun2022path}
Yann LeCun.
\newblock A path towards autonomous machine intelligence.
\newblock OpenReview, 2022.
\newblock URL \url{https://openreview.net/forum?id=BZ5a1r-kVsf}.

\bibitem[Ledoit \& Wolf(2004)Ledoit and Wolf]{ledoit2004}
Olivier Ledoit and Michael Wolf.
\newblock A well-conditioned estimator for large-dimensional covariance
  matrices.
\newblock \emph{Journal of Multivariate Analysis}, 88\penalty0 (2):\penalty0
  365--411, 2004.
\newblock \doi{10.1016/S0047-259X(03)00096-4}.

\bibitem[Li et~al.(2021)Li, Kovachki, Azizzadenesheli, Liu, Bhattacharya,
  Stuart, and Anandkumar]{li2021fno}
Zongyi Li, Nikola~Borislavov Kovachki, Kamyar Azizzadenesheli, Burigede Liu,
  Kaushik Bhattacharya, Andrew Stuart, and Anima Anandkumar.
\newblock Fourier neural operator for parametric partial differential
  equations.
\newblock In \emph{International Conference on Learning Representations}, 2021.

\bibitem[Liu \& Tegmark(2021)Liu and Tegmark]{liu2021conservation}
Ziming Liu and Max Tegmark.
\newblock Machine learning conservation laws from trajectories.
\newblock \emph{Physical Review Letters}, 126\penalty0 (18):\penalty0 180604,
  2021.
\newblock \doi{10.1103/PhysRevLett.126.180604}.

\bibitem[Locatello et~al.(2019)Locatello, Bauer, Lucic, Raetsch, Gelly,
  Sch{\"o}lkopf, and Bachem]{locatello2019challenging}
Francesco Locatello, Stefan Bauer, Mario Lucic, Gunnar Raetsch, Sylvain Gelly,
  Bernhard Sch{\"o}lkopf, and Olivier Bachem.
\newblock Challenging common assumptions in the unsupervised learning of
  disentangled representations.
\newblock In \emph{Proceedings of the 36th International Conference on Machine
  Learning}, volume~97 of \emph{Proceedings of Machine Learning Research}, pp.\
   4114--4124. PMLR, 2019.

\bibitem[Mialon et~al.(2023)Mialon, Garrido, Lawrence, Rehman, LeCun, and
  Kiani]{mialon2023lie}
Gr{\'e}goire Mialon, Quentin Garrido, Hannah Lawrence, Danyal Rehman, Yann
  LeCun, and Bobak Kiani.
\newblock Self-supervised learning with {Lie} symmetries for partial
  differential equations.
\newblock In \emph{Advances in Neural Information Processing Systems},
  volume~36, pp.\  28973--29004, 2023.

\bibitem[Odena et~al.(2016)Odena, Dumoulin, and Olah]{odena2016deconv}
Augustus Odena, Vincent Dumoulin, and Chris Olah.
\newblock Deconvolution and checkerboard artifacts.
\newblock \emph{Distill}, 1\penalty0 (10), 2016.
\newblock \doi{10.23915/distill.00003}.

\bibitem[Okubo(1970)]{okubo1970}
Akira Okubo.
\newblock Horizontal dispersion of floatable particles in the vicinity of
  velocity singularities such as convergences.
\newblock \emph{Deep Sea Research and Oceanographic Abstracts}, 17\penalty0
  (3):\penalty0 445--454, 1970.
\newblock \doi{10.1016/0011-7471(70)90059-8}.

\bibitem[Perez et~al.(2018)Perez, Strub, de~Vries, Dumoulin, and
  Courville]{perez2018film}
Ethan Perez, Florian Strub, Harm de~Vries, Vincent Dumoulin, and Aaron
  Courville.
\newblock {FiLM}: Visual reasoning with a general conditioning layer.
\newblock \emph{Proceedings of the AAAI Conference on Artificial Intelligence},
  32\penalty0 (1), 2018.
\newblock \doi{10.1609/aaai.v32i1.11671}.

\bibitem[Piloto et~al.(2022)Piloto, Weinstein, Battaglia, and
  Botvinick]{piloto2022intuitive}
Luis~S. Piloto, Ari Weinstein, Peter Battaglia, and Matthew Botvinick.
\newblock Intuitive physics learning in a deep-learning model inspired by
  developmental psychology.
\newblock \emph{Nature Human Behaviour}, 6\penalty0 (9):\penalty0 1257--1267,
  2022.
\newblock \doi{10.1038/s41562-022-01394-8}.

\bibitem[Sanchez-Gonzalez et~al.(2020)Sanchez-Gonzalez, Godwin, Pfaff, Ying,
  Leskovec, and Battaglia]{sanchezgonzalez2020gns}
Alvaro Sanchez-Gonzalez, Jonathan Godwin, Tobias Pfaff, Rex Ying, Jure
  Leskovec, and Peter Battaglia.
\newblock Learning to simulate complex physics with graph networks.
\newblock In \emph{Proceedings of the 37th International Conference on Machine
  Learning}, volume 119 of \emph{Proceedings of Machine Learning Research},
  pp.\  8459--8468. PMLR, 2020.

\bibitem[Shi et~al.(2016)Shi, Caballero, Husz{\'a}r, Totz, Aitken, Bishop,
  Rueckert, and Wang]{shi2016subpixel}
Wenzhe Shi, Jose Caballero, Ferenc Husz{\'a}r, Johannes Totz, Andrew~P. Aitken,
  Rob Bishop, Daniel Rueckert, and Zehan Wang.
\newblock Real-time single image and video super-resolution using an efficient
  sub-pixel convolutional neural network.
\newblock In \emph{Proceedings of the IEEE Conference on Computer Vision and
  Pattern Recognition (CVPR)}, pp.\  1874--1883, 2016.
\newblock \doi{10.1109/CVPR.2016.207}.

\bibitem[Weiss(1991)]{weiss1991}
John Weiss.
\newblock The dynamics of enstrophy transfer in two-dimensional hydrodynamics.
\newblock \emph{Physica D: Nonlinear Phenomena}, 48\penalty0 (2--3):\penalty0
  273--294, 1991.
\newblock \doi{10.1016/0167-2789(91)90088-Q}.

\bibitem[Wiskott \& Sejnowski(2002)Wiskott and Sejnowski]{wiskott2002sfa}
Laurenz Wiskott and Terrence~J. Sejnowski.
\newblock Slow feature analysis: Unsupervised learning of invariances.
\newblock \emph{Neural Computation}, 14\penalty0 (4):\penalty0 715--770, 2002.
\newblock \doi{10.1162/089976602317318938}.

\end{thebibliography}
\bibliographystyle{tmlr}

\clearpage
\appendix
\phantomsection\label{startofappendix}
\counterwithin*{table}{section}\renewcommand{\thetable}{\thesection\arabic{table}}\renewcommand{\theHtable}{\thesection.\arabic{table}}
\counterwithin*{figure}{section}\renewcommand{\thefigure}{\thesection\arabic{figure}}\renewcommand{\theHfigure}{\thesection.\arabic{figure}}
\section{Study 1: every model under the pre-registered measures}\label{app:A}

\begin{table}[htbp]
\centering
\footnotesize
\caption{Means over 3 seeds. Probe rows are $R^2$ after partialling out time $\times$ initial-condition family, except the $\nu$ row, which is not partialled; S3 rows are $|r|$. Seed standard deviations, the shift-averaged readout and MLP probes are in \texttt{REPORT.md}, appendix A1--A2. Applying the full pre-registered rules to every model against the control, only static VICReg meets S1 (4 of 5, including Okubo--Weiss), and only the pixel predictor meets S3; no model meets S2.}
\label{tab:A1}
\begin{tabularx}{\linewidth}{>{\raggedright\arraybackslash}Xcccccc}
\toprule
Quantity (in-distribution test, raw pooled latent) & Untrained & JEPA & Pixel & Pooled JEPA & Static VICReg & Non-physical JEPA \\
\midrule
log enstrophy, $R^2$ & 0.727 & 0.930 & 0.890 & 0.862 & 0.973 & 0.956 \\
log palinstrophy, $R^2$ & 0.437 & 0.746 & 0.733 & 0.665 & 0.900 & 0.802 \\
log(1 + vortex count), $R^2$ & 0.531 & 0.752 & 0.521 & 0.637 & 0.842 & 0.789 \\
Okubo--Weiss fraction, $R^2$ & 0.456 & 0.507 & 0.270 & 0.423 & 0.673 & 0.504 \\
log dissipation, $R^2$ & 0.134 & 0.176 & 0.156 & 0.148 & 0.191 & 0.178 \\
$\log\nu$, frame pair, $R^2$ & 0.481 & 0.457 & 0.710 & 0.317 & 0.699 & 0.522 \\
S3: enstrophy, best slow-direction $|r|$ & 0.514 & 0.356 & 0.821 & 0.431 & 0.486 & 0.470 \\
S3: energy, best slow-direction $|r|$ & 0.948 & 0.794 & 0.916 & 0.702 & 0.667 & 0.734 \\
\bottomrule
\end{tabularx}
\end{table}

\section{Exploratory line between Studies 1 and 2}\label{app:B}
\begin{figure}[htbp]
\centering
\includegraphics[width=0.62\linewidth]{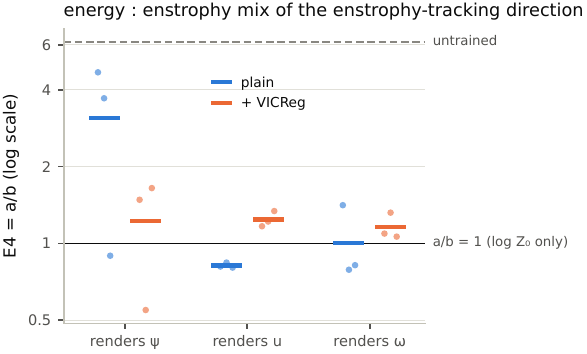}
\caption{E4~$= a/b$, the energy--enstrophy mix of the enstrophy-tracking direction (Section~\ref{sec:5.2}), for the plain and VICReg autoencoders of Figure~\ref{fig:2} on the exploration set: seeds and seed mean, log axis. E4~$= 1$ means the direction tracks $\log Z_0$ alone; the dashed line is the untrained control (E).}
\label{fig:B1}
\end{figure}

\subsection{From one unplanned result to a hypothesis (exploratory)}\label{sec:5.1}

Study 1's only physics-like positive was unplanned: the pixel predictor's enstrophy-tracking slow direction. We followed it on fresh data, with the Study 1 test sets retired. The exploration set held 1,006 trajectories (378 inviscid). A matched-energy set held 300 inviscid trajectories that all start at the same energy ($E_0$ = 0.2893), so that an enstrophy direction cannot ride on energy. A kill gate, committed before any computation, required the lead to survive on both. It did: E1 was 0.829 against 0.418 for the untrained control on the exploration set, and 0.979 against 0.177 at matched energy (E).

The follow-up stages then removed the explanation we had started with. Prediction horizon barely mattered: across $k = 1$, 4, 16 and 32 frames, the pixel predictor (``imaginer'') scored E1 0.68--0.80 and a JEPA at the same horizons (``summarizer'') 0.31--0.46. The imaginer showed no trend in k (its $k = 1$ point rests on two seeds and its $k = 16$ point on one); the summarizer rose slowly with k (0.31, 0.37, 0.38, 0.46). Predicting the future was not needed at all: plain autoencoders that render the present frame reached the imaginer's level when rendering vorticity (0.81) or velocity (0.82), and less when rendering the streamfunction (0.53). Adding VICReg's variance and covariance terms to those autoencoders dropped them to around control level (0.52, 0.42 and 0.33 against 0.42), while raising linear decodability, for example log palinstrophy $R^2$ from 0.68 to 0.93 for the vorticity autoencoder. Finally, shifting the imaginer's ($k = 4$) context maps $\pm 2$ standard deviations along its enstrophy direction changed the rendered field's enstrophy by only about 1--5\%, within the 95th percentile of random directions in two of three seeds, and with energy and palinstrophy moving together: a first sign that a prominent direction need not be one the decoder uses (E).

\begin{table}[htbp]
\centering
\small
\caption{Exploratory (E). Horizon is confounded with training length, and the imaginer's $k = 1$ and $k = 16$ cells rest on two seeds and one seed.}
\label{tab:B1}
\begin{tabularx}{\linewidth}{>{\raggedright\arraybackslash}X>{\raggedright\arraybackslash}X}
\toprule
Item & Value \\
\midrule
Exploration data & 1,006 trajectories (378 inviscid); matched-energy set of 300 inviscid trajectories at $E_0$ = 0.2893 \\
JEPA enstrophy $|r|$ by readout: pooled / map mean $\oplus$ std / map PCA-512 & 0.353 / 0.303 / 0.295 \\
Kill gate (a), exploration set: pixel vs untrained & 0.829 vs 0.418, gap $+0.411$ [$+0.326$, $+0.514$]: pass \\
Kill gate (b), matched energy: pixel vs untrained & 0.979 vs 0.177, gap $+0.802$ [$+0.718$, $+0.827$]: pass \\
E1 by horizon, imaginer / summarizer ($k = 1$, 4, 16, 32) & 0.68 (two seeds) / 0.31, 0.80 / 0.37, 0.69 (one seed) / 0.38, 0.76 / 0.46 \\
Imaginer $-$ summarizer gap by k & $+0.37$, $+0.42$, $+0.31$, $+0.30$ (all intervals above zero) \\
Autoencoder E1, plain: renders $\psi$ / $u$ / $\omega$ & 0.53 / 0.82 / 0.81 \\
Autoencoder E1, + VICReg: renders $\psi$ / $u$ / $\omega$ & 0.33 / 0.42 / 0.52 \\
Untrained control E1 (exploration set) & 0.42 \\
$\log P R^2$, vorticity autoencoder plain $\rightarrow$ + VICReg & 0.68 $\rightarrow$ 0.93 \\
Rendered enstrophy change at $\pm 2$ SD along the enstrophy direction & about 1--5\% \\
Compute & 24.5 logged hours \\
\bottomrule
\end{tabularx}
\end{table}

\section{The dial build, stage by stage}\label{app:C}

This design was reached adaptively. A first model with a shared trunk and two render heads produced no dial in either variant; in the variant with a FiLM code, normalized content still carried loudness ($R^2$ 0.99), and the code's loudness direction failed the round trip. Separate branches with scale-free content produced a loudness dial (later withdrawn by the artefact criterion) but no shape dial: the dial code encoded $s$ at $R^2$ 0.95, yet steering it did not move $s$, consistent with the content branch still supplying shape. Fully whitening the content spectrum forced shape into the dial code and produced a shape dial in some seeds, but made rendering 3.8 times worse than the VICReg autoencoder. The shape-noise dose replaced full whitening, and the decoder change came last (Section~\ref{sec:6.4}).

\textbf{Latent orthogonality is not output independence.} Our first shape rule projected the loudness direction out of the dial code and took the leading remaining principal component. It returned 50/50 mixtures in every seed ($\sigma_{\mathrm{shape}}$ 0.50--0.54): the direction was exactly orthogonal to loudness in the latent, yet steering it changed loudness about as much as shape. Orthogonality in the latent's $z$-scored coordinates says nothing about the decoder's response, so independence has to be imposed on the output.

A third rule, the singular value decomposition of the output spectrum's response to the dial code, failed: it ranks directions by raw sensitivity, so it found directions along which the data barely vary.

\begin{figure}[htbp]
\centering
\includegraphics[width=\linewidth]{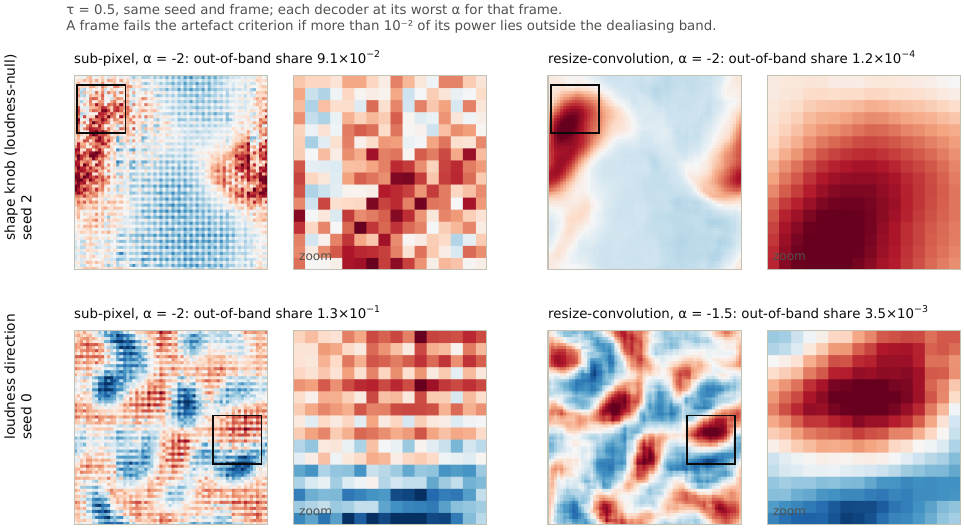}
\caption{Steered fields at $\tau = 0.5$ from a sub-pixel decoder and from the resize-convolution decoder: same seed and frame, each at its worst steering amplitude, with the out-of-band power share above each panel (E).}
\label{fig:5}
\end{figure}

\begin{figure}[htbp]
\centering
\includegraphics[width=\linewidth]{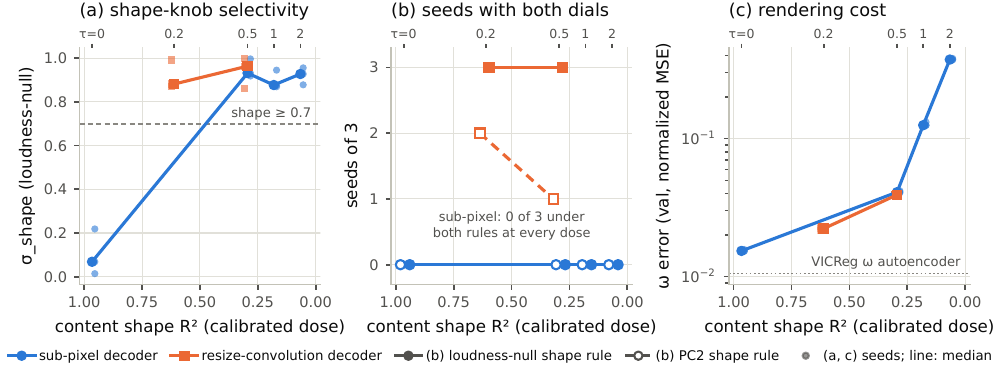}
\caption{As the content's shape signal is degraded, the loudness-null direction becomes a shape knob; both dials pass only with the resize decoder; rendering error rises with dose. The curve mixes decoders: the $\tau = 0.2$ point and the second $\tau = 0.5$ point use the resize-convolution decoder, the rest the sub-pixel decoder (E).}
\label{fig:6}
\end{figure}

\begin{table}[htbp]
\centering
\small
\caption{Exploratory (E). Several design and rule choices were made after seeing results; each stage's predictions were committed before it ran.}
\label{tab:C1}
\begin{tabularx}{\linewidth}{c>{\raggedright\arraybackslash}X>{\raggedright\arraybackslash}X}
\toprule
Stage & What changed & What it showed \\
\midrule
0 & Measure 21 existing models & 0 of 84 directions are dials (best effect 0.086); the supervised $s$-direction is readable ($|r|$ 0.97--0.99) but moves nothing \\
1 & Channel-split latent, two render heads, FiLM variant & No dials; normalized content still carries loudness ($R^2$ 0.99) \\
2 & Separate branches, scale-free content & Loudness dial in 3 of 3 seeds (later withdrawn by the artefact criterion); the dial code holds $s$ at $R^2$ 0.95 but steering it does not move $s$ \\
2b & Whitened content & PCA shape dial in 2 of 3 seeds (1 of 3 after the artefact criterion); no clean loudness dial; rendering 3.8 times the VICReg autoencoder \\
A & No training: loudness by symmetry; grid-artefact criterion & Pure loudness dial in 2 of 3 seeds; earlier loudness dials fail on tails (checkerboards in 5--14\% of frames at $\alpha = +2$) \\
A$'$ & No training: shape = top PC after projecting out loudness in the latent & 50/50 mixtures ($\sigma_{\mathrm{shape}}$ 0.50--0.54) \\
B & Shape noise on the content spectrum, $\tau = 0$, 0.5, 1, 2 & Loudness-null $\sigma_{\mathrm{shape}}$ rises from 0.07 to 0.93 as content shape $R^2$ falls; both dials 3 of 3 at $\tau = 0.5$ before the artefact criterion, 0 of 3 after; rendering 0.015 $\rightarrow$ 0.374 \\
C & $\tau = 0.5$ with resize-convolution decoder & 0.0\% artefact frames; both dials 3 of 3 (loudness-null), 1 of 3 (PC2); rendering 0.039 \\
D & $\tau = 0.2$ with resize-convolution decoder & Both dials 3 of 3 (loudness-null), 2 of 3 (PC2); rendering 0.022; selected for confirmation by a committed dose rule \\
\bottomrule
\end{tabularx}
\end{table}

\begin{table}[htbp]
\centering
\footnotesize
\caption{Final exploration model ($\tau = 0.2$, resize decoder), per seed (E). The sign of a direction is arbitrary.}
\label{tab:C2}
\begin{tabularx}{\linewidth}{c>{\raggedright\arraybackslash}Xc>{\raggedright\arraybackslash}Xcc}
\toprule
Seed & Shape dial: $\Delta s$ per unit $\alpha$, exploration / matched & $\sigma_{\mathrm{shape}}$ & Loudness dial: $\Delta a$ per unit $\alpha$ & $\sigma_{\mathrm{shape}}$ & Artefact frames \\
\midrule
0 & $+0.43$ / $+0.46$ & 0.99 / 0.95 & $+0.66$ / $+0.69$ & 0.00 / 0.03 & 0\% \\
1 & $+0.33$ / $+0.29$ & 0.87 / 0.82 & $+0.53$ / $+0.50$ & 0.00 / 0.02 & 0\% \\
2 & $+0.35$ / $+0.33$ & 0.88 / 0.86 & $-0.62$ / $-0.66$ & 0.00 / 0.05 & 0\% \\
\bottomrule
\end{tabularx}
\end{table}

\section{Study 3: secondary and descriptive results}\label{app:D}
\begin{figure}[htbp]
\centering
\includegraphics[height=0.86\textheight]{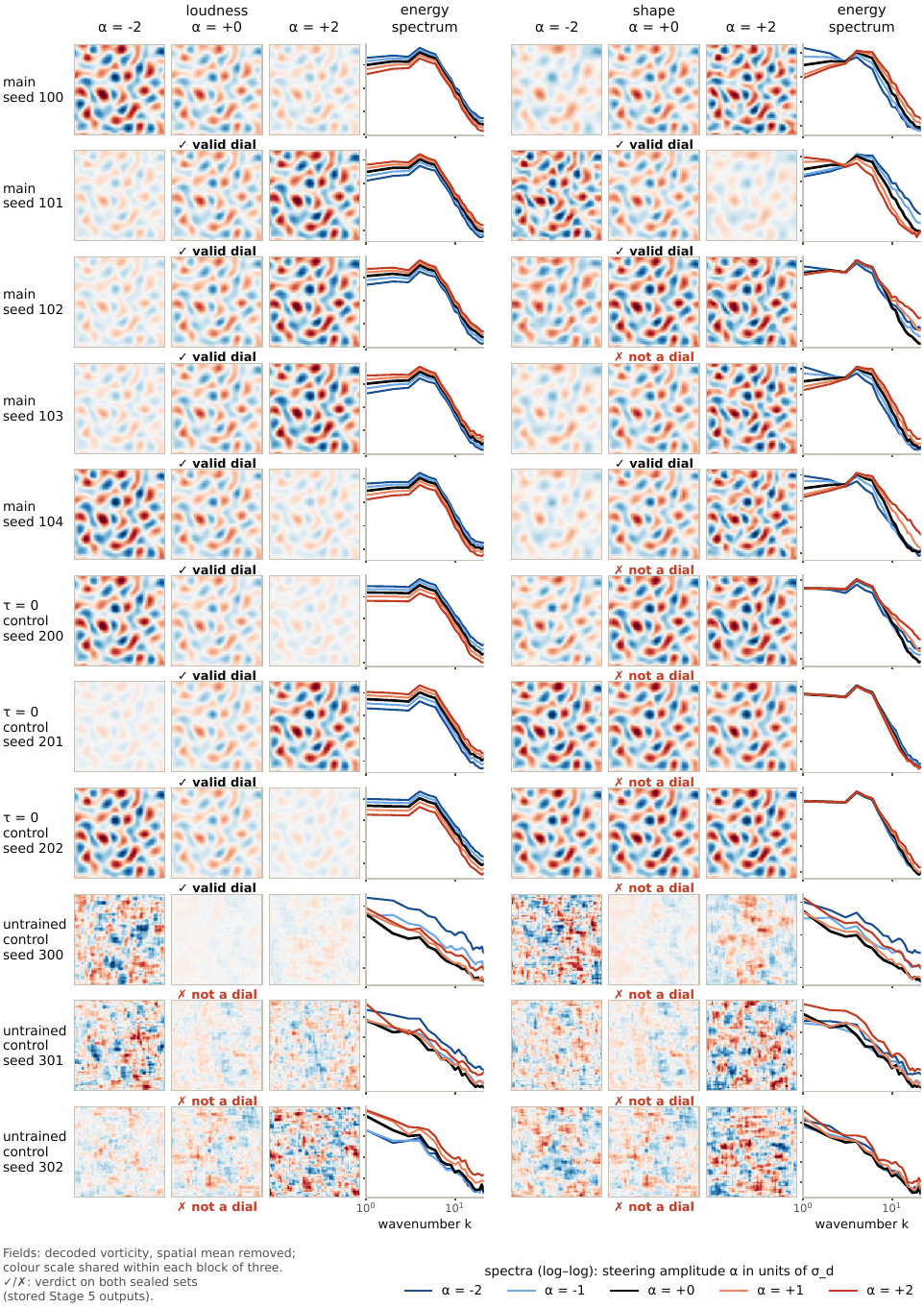}
\caption{As Figure~\ref{fig:7}, for all 11 scored models: main seeds 100--104, $\tau = 0$ seeds 200--202 and untrained seeds 300--302, decoded on the same exploration frame; \ding{51}/\ding{55} give each direction's verdict on both sealed sets (C).}
\label{fig:D1}
\end{figure}

\begin{table}[htbp]
\centering
\small
\caption{Study 3, pre-registered secondary and descriptive results (C), except where marked.}
\label{tab:D1}
\begin{tabularx}{\linewidth}{>{\raggedright\arraybackslash}X>{\raggedright\arraybackslash}X}
\toprule
Item & Value \\
\midrule
PC2 shape rule: seeds with both dials & 3 of 5 (100, 102, 104) \\
PC2 failures & 101: $\sigma_{\mathrm{shape}}$ 0.63 on the general set; 103: $\sigma_{\mathrm{shape}}$ 0.36 / 0.30 (mostly loudness) \\
$|\cos|$ between the loudness-null and PC2 directions, seeds 100--104 & 0.99, 0.95, 0.35, 0.87, 0.96 (computed after the run) \\
$\sigma_d$, loudness-null / PC2, seeds 100--104 & 5.12 / 5.12, 5.22 / 5.16, 2.11 / 4.47, 5.34 / 5.05, 3.70 / 3.84 \\
$\tau = 0$ control: loudness dial; loudness-null and PC2 effects & 3 of 3 (effect 0.78--0.85, $\sigma_{\mathrm{shape}} \leq$ 0.01); 0.02--0.10 and 0.12--0.23, all failing the cycle check \\
Untrained control & no dials; artefacts in 100\% of frames; rendering error 1.9--16.2 \\
Rendering (normalized $\omega$ error): main / $\tau = 0$ / VICReg autoencoder & 0.021--0.023 / 0.014--0.016 / 0.0097--0.0105 \\
Spatial-mean offset (> 10\% of $\omega^2$ in the mean): main loudness dials / seed 103 shape dial at $\alpha = -2$ / untrained & $\leq$ 2.3\% / 23.0\% of general-set frames / 41--100\% \\
Calibration probabilities in the pre-registration (written by the coding agent) & pass 60\%, partial 20\%, fail 20\% $\rightarrow$ partial; $\tau = 0$ within rule 85\% $\rightarrow$ yes; untrained within rule 90\% $\rightarrow$ yes \\
Compute & training 6.19 GPU-hours; about 0.7 lost to outages and a pause; evaluation 1.94; total about 8.8 of 10 \\
\bottomrule
\end{tabularx}
\end{table}

\section{Post hoc analysis of the Study 3 failures}\label{app:E}

\subsection{Why seeds 102 and 104 failed (post hoc)}\label{sec:6.6}

This analysis was chosen after the confirmation and run on exploration data only; it does not change the verdict (PH).

In seed 102, shape is not hidden along the loudness gradient: the gradient $g$ is orthogonal to the PC2 shape dial in the $z$-scored dial space ($\cos^2$ 0.000), and projecting $g$ out removes none of PC2's effect. What differs is the variance ranking after z-scoring. In the loudness-null subspace, the leading direction has variance 2.39 while PC2's projection has 1.48; in the other four seeds the two are within 0.15. The rule therefore picked a direction with less than half of PC2's spread ($\sigma_d$ 2.11 against 4.47), which steered weakly and non-monotonically. That z-scoring causes this is inferred from the variance table, not tested. In seed 104 the chosen direction is nearly PC2 ($|\cos|$ 0.96) and misses only the envelope tolerance on matched-energy data; the same miss ($-0.09$) appears on the exploration matched-energy set, so it belongs to the direction rather than to chance in the sealed set.

The two shape rules fail on different seeds. Counted together, a valid shape dial exists in all five main seeds, while the union finds none in any of the six controls. This reading was not pre-registered, it gives each seed two attempts, and it counts for nothing in the verdict. A rule designed after the failures, which picks the direction whose effect best matches the content-noise transformation itself, found no shape dial in any model: its directions were mostly shape ($\sigma_{\mathrm{shape}}$ 0.89--0.97 on the exploration set, 0.62--0.74 at matched energy) but too weak (effect 0.12--0.17 and cycle gain 0.35--0.49 on the exploration set), consistent with its objective rewarding a change that is well explained rather than one that is large. Its objective nevertheless separated the three groups here: about 0.10 for main models, 0.34--0.41 at $\tau = 0$ and 0.82--0.87 untrained.

\begin{table}[htbp]
\centering
\footnotesize
\caption{Post hoc (PH), exploration data only. $g$ is the mean gradient of output loudness with respect to the $z$-scored dial code. In seed 103, projecting out $g$ increases PC2's shape effect, because there PC2 mixes in loudness that works against shape.}
\label{tab:E1}
\begin{tabularx}{\linewidth}{cc>{\raggedright\arraybackslash}X>{\raggedright\arraybackslash}X>{\raggedright\arraybackslash}X}
\toprule
Seed & $\cos^2(g, \mathrm{PC2})$ & Fraction of PC2's shape effect removed by projecting out $g$ (linear / steered) & Variance after projection: top direction / along PC2 & $\sigma_d$, loudness-null / PC2 \\
\midrule
100 & 0.009 & $+0.07$ / $+0.08$ & 2.91 / 2.90 & 5.12 / 5.12 \\
101 & 0.057 & $+0.17$ / $+0.18$ & 2.44 / 2.43 & 5.22 / 5.16 \\
102 & 0.000 & $-0.00$ / $-0.00$ & 2.39 / 1.48 & 2.11 / 4.47 \\
103 & 0.117 & $-0.66$ / $-0.70$ & 2.32 / 2.17 & 5.34 / 5.05 \\
104 & 0.001 & $+0.02$ / $+0.01$ & 3.08 / 2.94 & 3.70 / 3.84 \\
\bottomrule
\end{tabularx}
\end{table}

The transformation-matched rule minimizes $J_T$, the fraction of a direction's steered change that the content-noise family (fixed-energy log-k tilt and curvature) cannot explain. It found no shape dial in 5 main and 6 control models. $J_T$ separated the groups (main 0.102--0.111, $\tau = 0$ 0.344--0.410, untrained 0.819--0.866) and flagged the failed seed-102 direction (0.69 against 0.19--0.35 for the other loudness-null directions), but not seed 104's (0.27). With the mean kept rather than removed, seed 103's classification was unchanged on the exploration sets; the pre-removal offset appeared in 16.4\% of explore-set frames at the worst $\alpha$ (0\% at matched energy). This analysis used 2.27 GPU-hours.

\section{Deviations and incidents}\label{app:F}

\begin{table}[htbp]
\centering
\small
\caption{Every entry is dated and tied to a commit in the released timeline (\texttt{results/paper/TIMELINE.md}).}
\label{tab:F1}
\begin{tabularx}{\linewidth}{l>{\raggedright\arraybackslash}X}
\toprule
When & Event \\
\midrule
Before Study 1 & About 70 CPU planning pilots, summarized in the log, not reproducible from the repository \\
Study 1 & Pre-registration committed before training; amendment A1 seven minutes later (no-collapse rule limited to collapse-prone objectives) \\
Study 1 & Validation run crashed at its final console print after writing complete outputs; output-only fix \\
Study 1 & Validation results seen before the test; afterwards only the encoding fix and an unscored traversal readout changed \\
Exploration & Stage 2 launch aborted by a line-ending bug before any training; restarted from scratch \\
Exploration & Budget raised four times (from 14 to 30 GPU-hours), each before the work it covered \\
Exploration & Imaginer-family divergences: 8 of 15 at $3\times10^{-3}$; all 15 rerun at $10^{-3}$ before evaluation, where 3 of 15 diverged and were excluded under a committed rule; 11 of 30 in total \\
Exploration & Plan revised before Stage 2 evaluation; the contrast probe was designed after the imagination probe and is labelled post hoc \\
Study 2 & Change agreed before unsealing: fresh seeds for H1/H2 and H4 added; superseded an earlier ``no retraining'' plan \\
Build & Stage 0 harness re-run three times with two added diagnostics; classifications identical \\
Build & Effect floor raised (0.1 $\rightarrow$ 0.2), envelope redefined, cycle gain added before Stage 1 training \\
Build & Start-up crash from a non-deterministic GPU median before any training step; median moved to CPU, runs repeated unchanged \\
Build & Grid-artefact criterion calibrated on one known-bad case, applied from then on and retroactively \\
Build & Shape rule changes: PC2 frozen as exploration primary, loudness-null added as secondary, effect-space SVD dropped \\
Build & Power outage during one run; rerun from scratch \\
Study 3 & Two outages and a pause before evaluation (about 0.7 GPU-hours lost); no partial checkpoint used \\
Study 3 & Rerunning seed 103 without mean removal on the sealed set was declined, because it would reopen a spent set; done on exploration data instead \\
\bottomrule
\end{tabularx}
\end{table}

The timeline also records five changes made before the work they governed: the build budget was raised from 15 to 18 and then 18.5 GPU-hours; the build's planned VICReg-on-dials dose curve was replaced by the content-side shape-noise dose; the build's rendering trigger was not applied in Study 3, as stated in advance; the exploration dropped the fake-physics and pixel-plus-VICReg models before evaluation, and the latter returned as H4; and Study 3's untrained control was given random FiLM initialization before pre-registration.

\subsection{Pre-registrations and sealed sets}\label{sec:8.1}

There were three pre-registrations, each committed before its test data were opened. The Study 1 pre-registration was committed before any model was trained; one amendment followed seven minutes later, after two one-epoch smoke runs and before any evaluation, restricting a no-collapse selection rule to objectives that can collapse (it does not apply to the pixel predictor). The Study 2 and Study 3 pre-registrations had no amendments. Every held-out set was opened exactly once by a frozen script: the Study 1 in-distribution and out-of-distribution test sets (in one run, single-use by protocol), and the code-sealed Study 2 confirmation set and two Study 3 sets. The Study 1 and Study 2 test sets were formally retired afterwards; the Study 3 sets are spent.

The committed Study 2 and Study 3 pre-registration files still carry the status line ``DRAFT for the investigator's review. It becomes binding when committed.'' Each file states that it binds on commit, the commit messages record the author's approval, and the seal code refused to open a sealed set unless the named pre-registration was committed and unmodified. We left the files as committed rather than edit them after the fact.

\subsection{Deviations that matter for interpretation}\label{sec:8.2}

\begin{itemize}
\item \textbf{Study 1.} Validation results were seen before the single test run. After that, only an output-encoding fix (a console print that crashed after writing complete outputs) and an unscored visualization readout were changed, and the final configuration was committed before the test run.
\item \textbf{Exploration.} At the learning rate chosen by the label-free selection rule ($3\times10^{-3}$), 8 of 15 imaginer-family runs diverged. Before any evaluation, every imaginer-family run was rerun at $10^{-3}$ and all $3\times10^{-3}$ runs were excluded. At $10^{-3}$, 3 of 15 diverged (one at $k = 1$, two at $k = 16$) and were excluded under a divergence rule committed before those runs (final validation loss at or above the copy-the-frame baseline), leaving two seeds at $k = 1$ and one at $k = 16.$ In total, 11 of 30 imaginer-family runs diverged. The plan was revised once before evaluation, and one probe designed after seeing another probe's result is labelled post hoc.
\item \textbf{Build.} The dial effect floor was raised from 0.1 to 0.2, the envelope test was redefined relative to an ideal dial, and a cycle-gain criterion was added, all after the first stage and before any training they governed. The grid-artefact criterion was calibrated on one known-bad case, then applied to every later result and retroactively, which withdrew earlier loudness dials. A start-up crash from a non-deterministic GPU median and one power outage led to reruns from scratch; no partial checkpoint was used.
\item \textbf{Study 3.} About 0.7 GPU-hours were lost to outages and a pause before evaluation. The eleven scored checkpoints were frozen by SHA-256 before the sealed sets were opened, and every checksum was verified at evaluation.
\end{itemize}

Log entries for the single test runs show the repository as modified (``+dirty''). In each case the only change was the run's own output files or the log line appended by unsealing; the tree was clean when the run started.

\subsection{Compute}\label{sec:8.3}

\begin{table}[htbp]
\centering
\small
\caption{Logged wall-clock hours on one NVIDIA RTX 3060 (12 GB), including CPU-bound phases within each line. About 70 CPU-only planning pilots before Study 1 are summarized in the log and not included.}
\label{tab:6}
\begin{tabularx}{\linewidth}{lr}
\toprule
Line of work & Logged hours \\
\midrule
Study 1 & 11.5 \\
Exploration and Study 2 & 29.6 \\
Dial build (exploration) & $\approx$ 17.8 \\
Study 3 confirmation & $\approx$ 8.8 \\
Post hoc analysis & 2.27 \\
\textbf{Total} & \textbf{$\approx$ 70} \\
\bottomrule
\end{tabularx}
\end{table}

\end{document}